\documentclass[journal]{IEEEtran}
\usepackage{array}
\usepackage[utf8x]{inputenc}
\usepackage[T1]{fontenc}
\usepackage[normalem]{ulem}
\usepackage{booktabs}
\usepackage{amsmath,amssymb,amsthm}
\usepackage{bbm}
\usepackage{lineno,hyperref}\hypersetup{colorlinks=true,hypertexnames=false}
\modulolinenumbers[5]
\usepackage{graphicx}
\usepackage{mwe}
\usepackage{graphbox}
\graphicspath{{./Fig/}}

\usepackage[usenames,dvipsnames]{xcolor}\definecolor{bg}{rgb}{0.95,0.95,0.95}
\usepackage[english]{babel}
\usepackage{listings}
\usepackage{multirow}
\usepackage{tabularx}

\usepackage{cite}

\usepackage[linesnumbered,ruled]{algorithm2e}

\usepackage{afterpage}

\usepackage{pgfplotstable}
\usepgfplotslibrary{fillbetween}
\usetikzlibrary{positioning,fit}
\usepgfplotslibrary{colorbrewer}
\pgfplotsset{cycle list/Dark2}
\pgfplotsset{compat=newest}
\usepackage{colortbl}
\pgfplotscreateplotcyclelist{mycolorlist}{%
blue\\%
magenta\\%
brown\\%
Emerald\\%
orange\\%
purple\\%
}
\usepackage{csvsimple}

\newcommand\norm[1]{\left\lVert#1\right\rVert}
\newcommand\prox[2]{\text{prox}_{#1}^{#2}}

\begin{document}
\title{Matrix cofactorization for joint spatial-spectral unmixing of hyperspectral images}

\author{Adrien~Lagrange,~\IEEEmembership{Student Member,~IEEE,}
        Mathieu~Fauvel,~\IEEEmembership{Senior Member,~IEEE,}\\
        St\'{e}phane~May
        and Nicolas~Dobigeon,~\IEEEmembership{Senior Member,~IEEE}
\thanks{A. Lagrange and N. Dobigeon are with University of Toulouse, IRIT/INP-ENSEEIHT, Toulouse, France (e-mail:\{adrien.lagrange, nicolas.dobigeon\}@enseeiht.fr).}
\thanks{M. Fauvel is with CESBIO, University of Toulouse, CNES/CNRS/INRA/IRD/UPS, Toulouse, France (e-mail:mathieu.fauvel@inra.fr).}
\thanks{S. May is with Centre National d'\'{E}tudes Spatiales (CNES), DCT/SI/AP, Toulouse, France (e-mail:stephane.may@cnes.fr).}
\thanks{N. Dobigeon is also with Institut Universitaire de France (IUF), France.}
\thanks{Part of this work has been supported by Centre National d'\'{E}tudes Spatiales (CNES), Occitanie Region, EU FP7 through the ERANETMED JC-WATER program (project ANR-15-NMED-0002-02 MapInvPlnt), the ANR-3IA Artificial and Natural Intelligence Toulouse Institute (ANITI), and the European Research Council under Grant
ERC FACTORY-CoG-681839.}}

 \markboth{IEEE TRANSACTIONS ON GEOSCIENCE AND REMOTE SENSING, VOL. XX, NO. XX, XX 2020}{LAGRANGE \emph{et al.}: atrix Cofactorization for Joint Spatial-Spectral Unmixing}

\maketitle

\begin{abstract}
Hyperspectral unmixing aims at identifying a set of elementary spectra and the corresponding mixture coefficients for each pixel of an image. As the elementary spectra correspond to the reflectance spectra of real materials, they are often very correlated yielding an ill-conditioned problem.
To enrich the model and to reduce ambiguity due to the high correlation, it is common to introduce spatial information to complement the spectral information. The most common way to introduce spatial information is to rely on a spatial regularization of the abundance maps. In this paper, instead of considering a simple but limited regularization process, spatial information is directly incorporated through the newly proposed context of spatial unmixing. Contextual features are extracted for each pixel and this additional set of observations is decomposed according to a linear model. Finally the spatial and spectral observations are unmixed jointly through a cofactorization model. In particular, this model introduces a coupling term used to identify clusters of shared spatial and spectral signatures. An evaluation of the proposed method is conducted on synthetic and real data and shows that results are accurate and also very meaningful since they describe both spatially and spectrally the various areas of the scene.
\end{abstract}

\begin{IEEEkeywords}
Image analysis, spectral unmixing, hyperspectral imaging, cofactorization.
\end{IEEEkeywords}

\IEEEpeerreviewmaketitle


\section{Introduction}
\label{sec:intro}

\IEEEPARstart{O}{ver} the past decades the huge potential of Earth observation has pushed the scientific community to develop automatic methods to extract information from the acquired data. Hyperspectral imaging is a specific image modality proposing a very rich information in the spectral domain. Each pixel is indeed a dense sampling of the reflectance spectrum of the underlying area, with usually hundreds of measurements from visible to infrared domains. The particularities of hyperspectral images have lead to the development of specific interpretation methods in order to fully benefit from this spectral information. Spectral unmixing methods~\cite{Bioucas-Dias2012} are in particular based on the assumption that the reflectance spectrum of a pixel is the result of the mixture of a reduced set of elementary spectra called endmembers. Each of these endmembers is the reflectance spectrum corresponding to a specific material present in the scene. An unmixing method aims at estimating the existing endmembers and at recovering the proportions of each material in a given pixel, collected in a so-called abundance vector. These abundance vectors allow, for example, the end-user to build abundance maps displaying the distribution of materials over the observed scene.

As hyperspectral images contain rich spectral information, many unmixing methods focus on exploiting it and often neglect spatial information. Many well-established methods process pixels without taking in consideration the basic idea that neighboring pixels are often very similar. The only shared information between pixels is a common endmember matrix~\cite{Bioucas-Dias2010,Thouvenin2015}. Nevertheless, advanced methods have been proposed to perform spatially informed spectral unmixing~\cite{Shi2014}. The most direct approach is to consider local spatial regularization of the abundance maps. Several works, such as SUnSAL-TV~\cite{Iordache2012} or S2WSU~\cite{Zhang2018}, proposed to use TV-norm regularization to achieve this goal. Identifying clusters of spectrally similar pixels, gathered in homogeneous groups, has been also used to impose spatial smoothing of the abundances, e.g., in~\cite{Wang2017,Eches2011,Eches2013}. In a different way, other works used the local neighborhood to identify the subset of endmembers present in the neighborhood. It is especially useful when dealing with a large number of endmembers~\cite{Canham2011,Deng2013}. Finally, at a lesser extent, the spatial information has also been used to help the extraction of endmembers. Indeed, endmember extraction is often performed before estimating the abundance vectors. Some preprocessing step were proposed to ease the extraction and the identification of pure pixels as the averaging of spectra over superpixels~\cite{Thompson2010} or the use of spatial homogeneity scalar factors~\cite{Zortea2009a}.

Overall it is noticeable that all these approaches tend to exploit the very simple idea that neighboring pixels should be spectrally similar. However, the spatial information contained in remote sensing images is richer than this simple statement. This work attempts to show that the conventional hyperspectral unmixing approach can leverage on the spatial information to help for spectral discrimination. It relies on the hypothesis that very spectrally similar pixels can be discriminated by analyzing their spatial contexts. For instance, vegetated areas generally lead to challenging unmixing tasks due to high correlations between signatures associated with distinct vegetation types. However, it may generally be easier to discriminate these types of vegetation by analyzing their respective spatial contexts, even extracted from a  gray-scale panchromatic image. As an example, hardwoods are expected to exhibit canopies different from conifers, resulting in different spatial textures. Similarly, crops are arranged with specific spatial patterns different from those characterizing grassland. Exploiting spatial patterns and textures descriptors is thus expected to be helpful to the unmixing process. To exploit this assumption, this paper proposes a model based on a cofactorization task to jointly infer common spatial and spectral signatures from the image.

Cofactorization methods, sometimes referred to as coupled dictionary learning, have been implemented with success in many application fields, e.g., for text mining~\cite{Wang2011}, music source separation~\cite{Yoo2010} and image analysis~\cite{Yokoya2012,Akhtar2018}, among others. The main idea is to define an optimization problem relying on two factorizing models supplemented by a coupling term enforcing a dependence between the two models. The method proposed in this article jointly considers a spectral unmixing model and a decomposition of contextual features computed from a panchromatic image of the same scene. The coupling term is interpreted as a clustering identifying groups of pixels sharing similar spectral signatures and spatial contexts. This method exhibits two major advantages: \textit{i)} it provides very competitive results even though the method is unsupervised (i.e., it estimates both endmember signatures and abundance maps) and \textit{ii)} it provides very insightful results since the scene is partitioned into areas characterized by spectral and spatial signatures.

The remaining of the article is organized as follows. Section~\ref{sec:pb-stat} defines the spectral and the spatial models and further discusses the joint cofactorization problem. Section~\ref{sec:optim} then details the optimization scheme developed to solve the resulting non-convex non-smooth minimization problem. An evaluation of the proposed joint model is then conducted first on synthetic data in Section~\ref{sec:synth-exp} and then on real data in Section~\ref{sec:real-exp}. Finally, Section~\ref{sec:ccl} concludes the paper and presents some research perspectives to this work.

\section{Towards spatial-spectral unmixing}
\label{sec:pb-stat}

The main goal of this section is to introduce a model capable of spectrally and spatially characterizing an hyperspectral image. In particular, instead of incorporating prior spatial information as a regularization~\cite{Iordache2012}, the concept of spatial unmixing, detailed in Section~\ref{sec:spa-mod}, is introduced alongside a conventional spectral unmixing model in order to propose a new joint framework of spatial-spectral unmixing.

\subsection{Spectral mixing model}
\label{sec:spe-mod}


Spectral unmixing aims at identifying the elementary spectra and the proportion of each material in a given pixel~\cite{Bioucas-Dias2012}. Each of the $P$ pixels $\mathbf{y}_p$ is a $d_1$-dimensional measurement of a reflectance spectrum and is assumed to be a combination of $R_1$ elementary spectra $\mathbf{m}_r$, called endmembers, with $R_1 \ll d_1$. The so-called abundance vector $\mathbf{a}_p \in \mathbb{R}^{R_1}$ refers to the corresponding mixing coefficients in this pixel.
In a general case, where no particular assumption is made on the observed scene, the conventional linear mixture model (LMM) is widely adopted to describe the mixing process. It assumes that the observed mixtures are linear combinations of the endmembers. Within an unsupervised framework, i.e., when both endmember signatures and abundances should be recovered, linear spectral unmixing can be formulated as the following minimization problem
\begin{equation} \label{eq:spe-mod}
  \min_{\mathbf{M},\mathbf{A}} \norm{\mathbf{Y} - \mathbf{M} \mathbf{A}}_{\mathrm{F}}^2 + \imath_{\mathbb{R}^{d_1 \times R_1}_+}{(\mathbf{M})} + \imath_{\mathbb{S}_{R_1}^P}(\mathbf{A})
\end{equation}
where the matrices $\mathbf{Y} \in \mathbb{R}^{d_1 \times P}$ gathers all the observed pixels, $\mathbf{M} \in \mathbb{R}^{d_1 \times R_1}$ the endmembers, $\mathbf{A} \in \mathbb{R}^{R_1 \times P}$ the abundance vectors and $\imath_{\mathbb{R}^{d_1 \times R_1}_+}{(\cdot)}$ and $\imath_{\mathbb{S}_{R_1}^P}(\cdot)$ are respectively  indicator functions on the non-negative quadrant and the column-wise indicator function on the $R_1$-dimensional probability simplex denoted by $\mathbb{S}_{R_1}$. The non-negative constraint over $\mathbf{M}$ is justified by the fact that endmember signatures are reflectance spectra and thus non-negative. The second indicator function enforces non-negative and sum-to-one constraints on the abundance vectors $\mathbf{a}_p$ ($p =1,\ldots,P$) in order to interpret them as proportion vectors. It is worth noting that the sum-to-one constraint is sometimes disregarded since it has been argued that relaxing this constraint out offers a better adaptation to possible changes of illumination in the scene~\cite{Drumetz2016}. Due to the usual ill-conditioning of the endmember matrix $\mathbf{M}$, the objective function underlying \eqref{eq:spe-mod} is often granted with additional regularizations promoting expected properties of the solution. In particular, numerous works exploited the expected spatial behavior of the mixing coefficients to introduce spatial regularizations enforcing piecewise-constant \cite{Eches2011, Iordache2012} or smoothly varying \cite{Thouvenin2015,Moussaoui2012whispers} abundance maps, possibly driven by external knowledge \cite{Uezato2018}. Conversely, this work does not consider spatial information as a prior knowledge but rather proposes a decomposition model dedicated to the image spatial content, paving the way towards the concept of \emph{spatial unmixing}. This contribution is detailed in the next sections.

\subsection{Spatial mixing model}
\label{sec:spa-mod}

As previously mentioned, this paper proposes to complement the conventional linear unmixing problem \eqref{eq:spe-mod} with an additional data-fitting term accounting for spatial information already contained in the hyperspectral image. To do so, for sake of generality, we assume that the scene of interest is characterized by vectors of spatial features $\mathbf{s}_p \in \mathbb{R}^{d_2}$ describing the context around the corresponding hyperspectral pixel indexed by $p$. The features can be extracted from the hyperspectral image directly or from any other available image of any modality of the same scene, with possibly better spatial resolution. A common choice for designing these features will be discussed later. To capture common spatial patterns, akin to a so-called \emph{spatial unmixing}, these $P$ $d_2$-dimensional spatial features vectors $\mathbf{s}_p$ gathered in a matrix $\mathbf{S} \in \mathbb{R}^{d_2 \times P}$ are linearly decomposed and recovered from the optimization problem
\begin{equation}
\label{eq:spat-model}
  \min_{\mathbf{D},\mathbf{U}} \norm{\mathbf{S} - \mathbf{D} \mathbf{U}}_{\mathrm{F}}^2 + \imath_{\mathbb{R}^{d_2 \times R_2}_+}{(\mathbf{D})} + \imath_{\mathbb{S}_{R_2}^P}(\mathbf{U})
\end{equation}
where $\mathbf{D}\in \mathbb{R}^{d_2 \times R_2}$ is a dictionary matrix and $\mathbf{U} \in \mathbb{R}^{R_2 \times P}$ the corresponding coding matrix. 

The spatial model underlying \eqref{eq:spat-model} can be interpreted as a dictionary-based representation learning task. It means that the image in the considered feature space can be decomposed as a sum of elementary patterns collected in the matrix $\mathbf{D}$ of spatial signatures. The corresponding coding coefficients are gathered in $\mathbf{U}$. The non-negativity constraints are imposed to ensure an additive decomposition similarly to what is done in the context of non-negative matrix factorization~\cite{Lee1999}. Finally, without any constraint on the norms of $\mathbf{U}$ and $\mathbf{D}$, the problem would suffer from a scaling ambiguity between $\mathbf{U}$ and $\mathbf{D}$. To cope with this issue, additional sum-to-one constraints are imposed on the columns of $\mathbf{U}$. This choice, which somehow brings some loss of generality when compared to normalizing the rows of $\mathbf{U}$, leads to possibly amplitude-varying atoms in $\mathbf{D}$. In other words, to describe similar spatial patterns of different amplitudes, additional
atoms should be included in the spatial dictionary $\mathbf{D}$. However, normalizing the columns in $\mathbf{U}$ has the advantage of leading to spatial and spectral representation vectors of same unit norm, which prevents any unbalancies in the coupling process introduced in the next section.

It is worth noting that a model similar to \eqref{eq:spat-model} was implicitly assumed in \cite{Vasudevan2015, Vasudevan2016, Vasudevan2018} where a single-band image acquired by scanning transmission electron microscopy is linearly unmixed by principal component analysis \cite{Jolliffe1986}, independent component analysis \cite{Hyvarinen2001}, N-FINDR \cite{Winter1999spie} or a deep convolutional neural networks. In these works, the spatial feature space is defined by the magnitude of a sliding 2D-discrete Fourier transform, which unlikely ensures the additivity, or at least linear separability, assumptions underlying the mixtures. As an alternative, the strategy adopted in this work relies on a patch-based representation of the image, as popularized by several seminal contributions in the literature. Combined with a linear decomposition underlying \eqref{eq:spat-model}, this representation can be easily motivated by the intrinsic property of image self-similarity. It has already shown its practical interest for various image processing tasks, including classification and denoising. As archetypal examples, patch-based dictionary learning methods leverage on linear models similar to \eqref{eq:spat-model} to capture spatial redundancies \cite{Aharon2006,Mairal2009,Dobigeon2010, Wei2015}. This self-similarity property is also a key assumption to motivate linear aggregation steps in many non-local denoising techniques such as NL-means \cite{Buades2005} and BM3D \cite{Dabov2007}. Its simplicity makes this model popular in many application fields, including medical imaging~\cite{Tong2013} and photography~\cite{Dong2011}. Thus, in the numerical experiments reported in Sections~\ref{sec:synth-exp} and~\ref{sec:real-exp}, the spatial features will be chosen as the elementary patches extracted from the virtual panchromatic image computed by averaging the hyperspectral bands.

\subsection{Coupling spatial and spectral mixing models}
\label{sec:coupling}
The two previous sections have defined two matrix factorization problems associated with two unmixing tasks considered independently. To mutually benefit from spectral and spatial information brought by the image, these two tasks should be considered jointly. A natural approach consists in coupling the two tasks by relating the coding factors involved in the two matrix decompositions. A hard coupling between the coding matrices, which would consist in constraining  \emph{spectral} and \emph{spatial} abundance maps to be equal (i.e., $\mathbf{A}=\mathbf{U}$), will be shown to be not sufficiently flexible to account for complex interactions between spatial and spectral information (see Section \ref{sec:synth-exp}). Conversely, the proposed approach relies on the simple yet sound assumption that the pixels in the scene obey a small number of spectral and spatial behaviors that can be clustered. This implicitly means that a given elementary material, uniquely represented by a single endmember spectrum, is expected to be present in the scene with a small number of distinct spatial configurations. 
Thus, the coupling is designed as a clustering task which aims at recovering common behaviors jointly exhibited by the spatial and spectral abundance maps. More precisely, this clustering task can be formulated as another nonnegative matrix factorization problem
\begin{align}
\label{eq:coupling-pb}
  \min_{\mathbf{B},\mathbf{Z}} & \norm{\left( {\begin{array}{c} \mathbf{A}\\ \mathbf{U}\\\end{array}} \right) - \mathbf{B} \mathbf{Z}}_{\mathrm{F}}^2 + \frac{\lambda_z}{2} \mathrm{Tr}(\mathbf{Z}^T\mathbf{V}\mathbf{Z})  \\
  &+ \imath_{\mathbb{R}^{(R_1+R_2)\times K}_+}{(\mathbf{B})} + \imath_{\mathbb{S}_{K}^P}(\mathbf{Z})
\end{align}
with $\mathbf{V} = \mathbf{1}_K \mathbf{1}_K^T - \mathbf{I}_K$ where $\mathbf{I}_K$ is the $K \times K$ identity matrix, $\mathbf{1}_K$ is the $K\times 1$ vector of ones and $\mathrm{Tr}(\cdot)$ is the trace operator. The two coding matrices $\mathbf{A}$ and $\mathbf{U}$ are concatenated and the clustering is  conducted on the columns of the resulting whole coding matrix. The matrix $\mathbf{Z} \in \mathbb{R}^{K \times P}$ describes the assignments to the clusters, where $\mathbf{z}_p$ gathers the probabilities of belonging to each of the clusters, hence the non-negativity and sum-to-one constraint enforced on it. It is accompanied with a specific regularization (see $2$nd term in \eqref{eq:coupling-pb}). This penalty promotes orthogonality over the lines of $\mathbf{Z}$ since it can be rewritten as $\mathrm{Tr}(\mathbf{Z}^T\mathbf{V}\mathbf{Z}) = \sum_{\substack{k_1\ne k_2}} \left<\mathbf{z}_{k_1,:}|\mathbf{z}_{k_2,:}\right>$ where $\left<\cdot|\cdot\right>$ stands for the scalar product. Due to non-negativity of the elements of $\mathbf{Z}$, this term becomes minimal when the assignments to clusters obey a hard decision, i.e., when one component of $\mathbf{z}_p$ is equal to $1$ and the others are set to $0$. Thus a strict orthogonality constraint would make the clustering problem \eqref{eq:coupling-pb} equivalent to a $k$-means problem~\cite{Pompili2014}. Centroids of the $K$ clusters define the columns of the matrix $\mathbf{B} \in \mathbb{R}^{(R_1 + R_2) \times K}$. Interestingly, each centroid is then the concatenation of a mean spectral abundance and a mean spatial abundance. In particular, it means that the pixels of a given cluster share the same spectral properties and a similar spatial context. Indeed, these centroids, when combined with the spectral and spatial signature matrices $\mathbf{M}$ and $\mathbf{D}$, provide a compact representation of the spectral and spatial contents of each cluster. More precisely, each column of the $(d_1+d_2) \times K$-matrix defined by 
\begin{equation}
\label{eq:mean_clusters}
 \left( \begin{array}{c} \bar{\mathbf{M}}\\ \bar{\mathbf{D}} \end{array} \right) = \left(
 \begin{array}{cc} \mathbf{M} & \boldsymbol{0}\\
 \boldsymbol{0} & \mathbf{D} 
 \end{array}
 \right) 
 \mathbf{B}
\end{equation}
can be interpreted as the spatial-spectral signature of each cluster, resulting from the concatenation of a mean spectral signature $\bar{\mathbf{m}}_k$ and a mean spatial signature $\bar{\mathbf{d}}_k$ ($k=1,\ldots,K$).

\subsection{Joint spatial-spectral unmixing problem}
Given the spectral mixing model recalled in Section \ref{sec:spe-mod}, the spatial mixing model introduced in Section \ref{sec:spa-mod} and their coupling term proposed in Section \ref{sec:coupling}, we propose to conduct spatial-spectral unmixing jointly by considering the overall minimization problem
\begin{align}
\label{eq:cofact-pb}
  \min_{\mathbf{M},\mathbf{A},\mathbf{D},\mathbf{U},\mathbf{B},\mathbf{Z}} &\frac{\lambda_0}{2} \norm{\mathbf{Y} - \mathbf{M} \mathbf{A}}_{\mathrm{F}}^2 + \imath_{\mathbb{R}^{d_1 \times R_1}_+}{(\mathbf{M})} + \imath_{\mathbb{S}_{R_1}^P}(\mathbf{A}) \nonumber\\
  &+ \frac{\lambda_1}{2} \norm{\mathbf{S} - \mathbf{D} \mathbf{U}}_{\mathrm{F}}^2 + \imath_{\mathbb{R}^{d_2 \times R_2}_+}{(\mathbf{D})} + \imath_{\mathbb{S}_{R_2}^P}(\mathbf{U}) \nonumber\\
  &+ \frac{\lambda_2}{2} \norm{\left( {\begin{array}{c} \mathbf{A}\\ \mathbf{U}\\\end{array}} \right) - \mathbf{B} \mathbf{Z}}_{\mathrm{F}}^2 + \frac{\lambda_z}{2} \mathrm{Tr}(\mathbf{Z}^T\mathbf{V}\mathbf{Z}) \nonumber\\
  &+ \imath_{\mathbb{R}^{(R_1+R_2)\times K}_+}{(\mathbf{B})} + \imath_{\mathbb{S}_{K}^P}(\mathbf{Z})
\end{align}
where $\lambda_0$, $\lambda_1$ and $\lambda_2$ adjust the respective contribution of the various fitting terms. It is worth noting that, because of the sum-to-one constraints enforced on the spectral abundance vectors $\mathbf{a}_p$ and spatial abundance vectors $\mathbf{u}_p$, all these coding vectors have the same unitary $\ell_1$-norm. It has the great advantage of avoiding a reweighing of $\mathbf{A}$ and $\mathbf{U}$ in the coupling term regardless of the number of endmembers and dictionary atoms. The next section describes the optimization scheme adopted to solve the joint spatial-spectral unmixing problem \eqref{eq:cofact-pb}, 

\section{Optimization scheme}
\label{sec:optim}

\subsection{PALM algorithm}

The cofactorization problem~\eqref{eq:cofact-pb} is a non-convex, non-smooth optimization problem. For these reasons, the problem remains very challenging to solve and requires the use of advanced optimization tools. The choice has been made to resort to the proximal alternating linearized minimization (PALM) algorithm~\cite{Bolte2014}. The core concept of PALM is to update each block of variables alternatively according to a proximal gradient descent step. This algorithm has the advantage of ensuring convergence to a critical point of the objective function even in the case of a non-convex, non-smooth problem.

In order to obtain these convergence results, the objective function has to ensure a specific set of properties. Firstly, the various terms of the objective function have to be separable in a sum of one smooth term $g(\cdot)$ and a set of independent non-smooth terms. Then, each of the independent non-smooth term has to be a proper, lower semi-continuous function $f_i: \mathbb{R}^{n_i} \rightarrow (-\infty,+\infty]$, where \(n_i\) is the input dimension of \(f_i\). Finally, a sufficient condition is that the smooth term is a $\mathcal{C}^2$-continuous function and that its partial gradients are globally Lipschitz with respect to the derivative variable. Further details are available in the original paper~\cite{Bolte2014}.

In problem~\eqref{eq:cofact-pb}, the smooth term $g(\cdot)$ is composed of the three quadratic terms and the orthogonality-promoting regularization. All these terms obviously verify the gradient Lipschitz and $\mathcal{C}^2$-continuous properties. Moreover, the non-smooth terms $f_i$ are separable into independent terms. Moreover, since they are all indicators functions on convex sets, their proximal operators are well-defined and, more specifically, are defined as the projection on the corresponding convex set. The projection on the non-negative quadrant is a simple thresholding of the negative values and the projection on the probability simplex can be achieved by a simple sort followed by a thresholding as described in~\cite{Condat2016}.

A summary of the overall optimization scheme is given in Algo.~\ref{alg:palm} where $L_\mathbf{X}$ stands for the Lipschitz constant of the gradient of $g(\cdot)$ considered as a function of $\mathbf{X}$. Partial gradients and Lipschitz moduli are all provided in Appendix~\ref{sec:app-optim}. Additional details regarding the implementation are discussed in what follows.

\begin{algorithm}[!ht]
\caption{PALM\label{alg:palm}}
\footnotesize
\SetAlgoLined
Initialize variables $\mathbf{M}^0$, $\mathbf{A}^0$, $\mathbf{D}^0$, $\mathbf{U}^0$, $\mathbf{B}^0$ and $\mathbf{Z}^0$\;
Set $\alpha > 1$\;
\While{\text{stopping criterion not reached}}{
  $\mathbf{M}^{k+1} \in \prox{\imath_{\mathbb{R}^{d_1\times R_1}_+}}{\alpha L_\mathbf{M}}(\mathbf{M}^{k} - \frac{1}{\alpha L_\mathbf{M}} \nabla_\mathbf{M}g(\mathbf{M}^k,\mathbf{A}^k,\mathbf{D}^k,\mathbf{U}^k,\mathbf{B}^k,\mathbf{Z}^k))$\;
  $\mathbf{A}^{k+1} \in \prox{\imath_{\mathbb{S}_{R_1}^P}}{\alpha L_\mathbf{A}}(\mathbf{A}^{k} - \frac{1}{\alpha L_\mathbf{A}} \nabla_\mathbf{A}g(\mathbf{M}^{k+1},\mathbf{A}^k,\mathbf{D}^k,\mathbf{U}^k,\mathbf{B}^k,\mathbf{Z}^k))$\;
  $\mathbf{D}^{k+1} \in \prox{\imath_{\mathbb{R}^{d_2 \times R_2}_+}}{\alpha L_\mathbf{D}}(\mathbf{D}^{k} - \frac{1}{\alpha L_\mathbf{D}} \nabla_\mathbf{D}g(\mathbf{M}^{k+1},\mathbf{A}^{k+1},\mathbf{D}^k,\mathbf{U}^k,\mathbf{B}^k,\mathbf{Z}^k))$\;
  $\mathbf{U}^{k+1} \in \prox{\imath_{\mathbb{S}_{R_2}^P}}{\alpha L_\mathbf{U}}(\mathbf{U}^{k} - \frac{1}{\alpha L_\mathbf{U}} \nabla_\mathbf{U}g(\mathbf{M}^{k+1},\mathbf{A}^{k+1},\mathbf{D}^{k+1},\mathbf{U}^k,\mathbf{B}^k,\mathbf{Z}^k))$\;
  $\mathbf{B}^{k+1} \in \prox{\imath_{\mathbb{R}^{(R_1+R_2)\times K}_+}}{\alpha L_\mathbf{B}}(\mathbf{B}^{k} - \frac{1}{\alpha L_\mathbf{B}} \nabla_\mathbf{B}g(\mathbf{M}^{k+1},\mathbf{A}^{k+1},\mathbf{D}^{k+1},\mathbf{U}^{k+1},\mathbf{B}^k,\mathbf{Z}^k))$\;
  $\mathbf{Z}^{k+1} \in \prox{\imath_{\mathbb{S}_{K}^P}}{\alpha L_\mathbf{Z}}(\mathbf{Z}^{k} - \frac{1}{\alpha L_\mathbf{Z}} \nabla_\mathbf{Z}g(\mathbf{M}^{k+1},\mathbf{A}^{k+1},\mathbf{D}^{k+1},\mathbf{U}^{k+1},\mathbf{B}^{k+1},\mathbf{Z}^k))$\;
}
\Return{$\mathbf{M}^{\textrm{end}},\mathbf{A}^{\textrm{end}},\mathbf{D}^{\textrm{end}},\mathbf{U}^{\textrm{end}},\mathbf{B}^{\textrm{end}},\mathbf{Z}^{\textrm{end}}$}
\end{algorithm}

\subsection{Implementation details}
\label{sec:impl}

\noindent{\textbf{Initialization and convergence --}} As explained, the PALM algorithm only ensures convergence to a critical point of the objective function. Hence, it is important to have a good initialization of the variables to be estimated. In the following experiments, the initial endmember matrix $\mathbf{M}^0$ has been chosen as the output of the vertex component analysis (VCA) \cite{Nascimento2005}. Abundance matrix is then initialized by solving the fully constrained least square problem $\min_{\mathbf{A}\in\mathbb{S}_{R_1}^{P}} \norm{\mathbf{Y}-\mathbf{MA}}_\mathrm{F}^2$. Finally, $\mathbf{D}^0$ and $\mathbf{U}^0$ are initialized by performing a $k$-means algorithm on columns of $\mathbf{S}$. Similarly $\mathbf{B}^0$ and $\mathbf{Z}^0$ are initialized by a $k$-means on the concatenation of $\mathbf{U}^0$ and $\mathbf{A}^0$.

As stated in Algo.~\ref{alg:palm}, a criterion is needed to monitor the convergence of the optimization algorithm. In the following experiments, the residual error of the objective function is computed at each iteration and, when the relative gap between the two last iterations is below a given threshold ($10^{-4}$ for these experiments), the algorithm is stopped.\\

\noindent{\textbf{Hyperparameters --}} Several weighting coefficient $\lambda_{\cdot}$ have been introduced in problem~\eqref{eq:cofact-pb} to adjust the respective contribution of each term. In the following experiments, some of these coefficients have been renormalized to take in consideration the respective dimensions and dynamics of the matrices defining each term, yielding
\begin{equation}
\label{eq:hyparameters}
  \left\{
  \begin{array}{rcl}
    \lambda_0 &=& \frac{1}{d_1\norm{\mathbf{Y}}_{\infty}^2} {\tilde{\lambda}_0}\\
    \lambda_1 &=& \frac{1}{d_2\norm{\mathbf{S}}_{\infty}^2} {\tilde{\lambda}_1}
  \end{array}
  \right..
\end{equation}

\section{Experiments using simulated data}
\label{sec:synth-exp}

Performance of the proposed spatial-spectral unmixing method has been assessed by conducting experiments on both synthetic and real data. The use of synthetic data makes quantitative validation possible whereas it is not possible with real data since there is no reference data.

\subsection{Data generation}
\label{sec:synth-data}

In order to properly evaluate the relevance of the proposed model, two synthetic images referred to as \textsf{\small Image 1} and \textsf{\small Image 2} have been generated such that they incorporate consistent spatial and spectral information. Note that the proposed simulation framework, detailed hereafter, does not directly rely on the forward model underlying the joint spectral-spatial unmixing problem in \eqref{eq:cofact-pb}.

\begin{figure}[ht]
  \centering
  \begin{tabular}{@{}c|cc}
    \textsf{\small Image 1} & \multicolumn{2}{c}{\textsf{\small Image 2}} \\ \includegraphics[width=0.24\columnwidth]{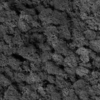}&                                                                         \includegraphics[width=0.24\columnwidth]{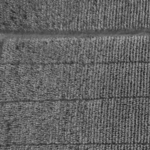}&                                                                              \includegraphics[width=0.24\columnwidth]{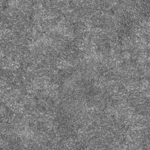}\\   
    \includegraphics[width=0.24\columnwidth]{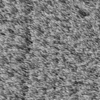}&
    \includegraphics[width=0.24\columnwidth]{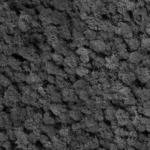}&
    \includegraphics[width=0.24\columnwidth]{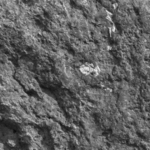}\\
    &\includegraphics[width=0.24\columnwidth]{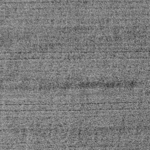}&
    \\
  \end{tabular}
  \caption{Synthetic dataset: textures (forest, wheat) for \textsf{\small Image 1} and (corn, grass, forest, rock, wheat) for \textsf{\small Image 2} (right).\label{fig:text}}
\end{figure}

\begin{figure}[ht]
  \centering
  \begin{tabular}{@{}cc@{~}c@{~}c@{~}c@{~}c}
    \rotatebox{90}{\textsf{\small Image 1}} &
    \includegraphics[width=0.17\columnwidth]{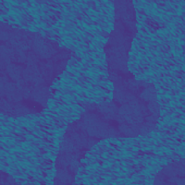}&
    \includegraphics[width=0.17\columnwidth]{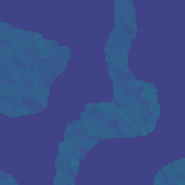}&
    \includegraphics[width=0.17\columnwidth]{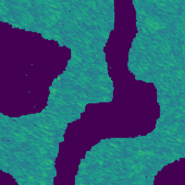}&
    \includegraphics[width=0.17\columnwidth]{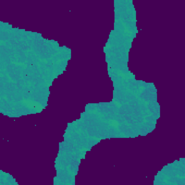}&\\
    &&&&\\
    \multirow{2}{*}{\rotatebox{90}{\textsf{\small Image 2}}} &
    \includegraphics[width=0.17\columnwidth]{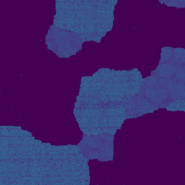}&
    \includegraphics[width=0.17\columnwidth]{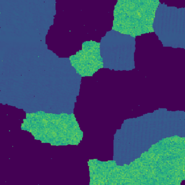}&
    \includegraphics[width=0.17\columnwidth]{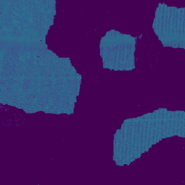}&
    \includegraphics[width=0.17\columnwidth]{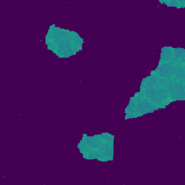}&
    \includegraphics[width=0.17\columnwidth]{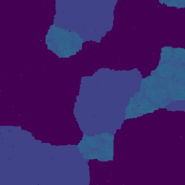}\\
    &
    \includegraphics[width=0.17\columnwidth]{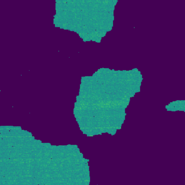}&
    \includegraphics[width=0.17\columnwidth]{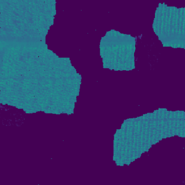}&
    \includegraphics[width=0.17\columnwidth]{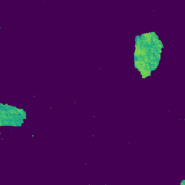}&
    \includegraphics[width=0.17\columnwidth]{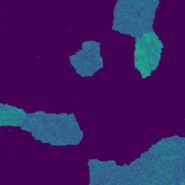}&\\
  \end{tabular}
  \caption{Synthetic dataset: abundance maps.\label{fig:synth-abund}}
\end{figure}

Each image is supposed to be composed of $J$ homogeneous regions with common spatial and spectral characteristics. These regions have been randomly generated according to a Potts-Markov random field with $J$ classes~\cite{Li2009}. Each region is characterized by specific (yet non piece-wise constant) spatial and spectral contents whose generation processes are described in what follows. According to the linear mixing model, the observed pixel spectra in each region are assumed to result from the linear combination of $R_1$ endmembers gathered in the common matrix $M=\left[\mathbf{m}_1,\ldots,\mathbf{m}_{R_1}\right]$
      \begin{equation}
       \forall p\in \mathcal{P}_j, \ \mathbf{y}_p = \mathbf{M}\mathbf{a}_p^{(j)}
      \end{equation}
      where $\mathcal{P}_j$ denotes the set of pixels belonging to the $j$th region and $\mathbf{a}_p^{(j)} \in \mathbb{S}_{R_1}$ is the abundance vectors  of the $p$th pixel in the $j$th region. The endmember signatures in $M$ have been extracted from the ASTER library~\cite{aster}.  Each of these regions is characterized by a particular texture. To ensure realistic spatial patterns, the $J$ textures associated with the $J$ regions are extracted from real panchromatic images. They are depicted in Figure~\ref{fig:text} for \textsf{\small Image 1} and \textsf{\small Image 2}. The intensity of the $p$th pixel of the grayscale image texture associated with the $j$th region is denoted ${t}_p^{(j)} \in (0,1)$ ($p \in\mathcal{P}_j$).  The key ingredient is the appropriate design of the abundance vectors $\mathbf{a}_p^{(j)}$ ($p\in\mathcal{P}_j$) which jointly encode the  spatial and spectral content in the $j$th region.  To include consistent spatial information, these abundance vectors are assumed to be convex combinations of two  pre-defined extreme spectral behaviors $\boldsymbol{\psi}_i^{(j)} \in \mathbb{S}_{R_1}$ ($i\in \left\{1,2\right\}$), i.e.,
      \begin{equation}
         \mathbf{a}_p^{(j)} = {t}_p^{(j)}\boldsymbol{\psi}_1^{(j)} + (1-{t}_p^{(j)})\boldsymbol{\psi}_2^{(j)}
      \end{equation}    
      where the grayscale intensity ${t}_p^{(j)} \in (0,1)$ of the $j$th texture modulates the spectral content in the $p$th pixel of the $j$th region. This supports the idea that a texture can be seen as small spatial variations of the proportions of elementary components between extreme spectral signatures. In particular, a white pixel in the $j$th texture (i.e., ${t}_p^{(j)}=1$) leads to a spectral content only driven by $\boldsymbol{\psi}_1^{(j)}$ and the corresponding pixel in $j$th region of the generated hyperspectral image is $\mathbf{y}_p = \mathbf{M}\boldsymbol{\psi}_1^{(j)}$. Conversely, a black pixel in the $j$th texture (i.e., ${t}_p^{(j)}=0$) leads to a spectral content only driven by $\boldsymbol{\psi}_2^{(j)}$ and the corresponding pixel is $\mathbf{y}_p = \mathbf{M}\boldsymbol{\psi}_2^{(j)}$. Of course, gray pixels in the texture lead to mixed behaviors with $\mathbf{y}_p = {t}_p^{(j)} \mathbf{M}\boldsymbol{\psi}_1^{(j)} + (1-{t}_p^{(j)})\mathbf{M}\boldsymbol{\psi}_2^{(j)}$. The generated abundance maps are shown in Figure~\ref{fig:synth-abund}. Again, note that this simulation protocol does not rely on the cofactorization formalism underlying the proposed algorithm.

\begin{figure}[!ht]
  \centering
  \begin{tabular}{@{}m{1em}c@{~}c@{~}c}
    \rotatebox{90}{\textsf{\small Image 1}} &
    \includegraphics[align=c,width=0.24\columnwidth]{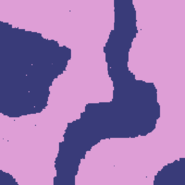}&
    \includegraphics[align=c,width=0.24\columnwidth]{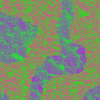}&
    \includegraphics[align=c,width=0.24\columnwidth]{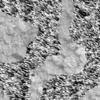}\\
    \rotatebox{90}{\textsf{\small Image 2}} &
    \includegraphics[align=c,width=0.24\columnwidth]{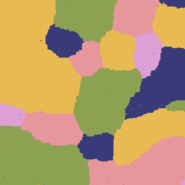}&
    \includegraphics[align=c,width=0.24\columnwidth]{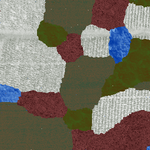}&
    \includegraphics[align=c,width=0.24\columnwidth]{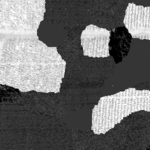}\\
    & (a) & (b) &(c) \\
  \end{tabular}
  \caption{Synthetic dataset: (a) segmentation map, (b) color composition of the hyperspectral image, (c) panchromatic image.\label{fig:data-synth}}
\end{figure}

Two images have been generated according to this process. \textsf{\small Image~1} is a $200 \times 200$-pixel image with $385$ spectral bands composed of $R_1=4$ endmembers and $J=2$ regions.  \textsf{\small Image~2} is a $300\times 300$-pixel image with $385$ spectral bands with $R_1=9$ endmembers  and $J=5$ regions. Note that for these two images, the texture-based modulating intensity ${t}_p^{(j)}$ never reaches the extreme values $0$ and $1$ ($\forall p, \forall j$). Additionally, from the two hyperspectral images, corresponding panchromatic images have been generated by first dividing each band individually by its empirical mean over the pixels and then summing all these normalized bands for each pixel. The contrast has been finally adjusted such that the mininum and maximum values of the image are 0 and 255, respectively, to obtain a 8-bit image. The generated hyperspectral and panchromatic images are shown in Figure~\ref{fig:data-synth}.

\subsection{Compared methods}
\label{sec:comp-meth}
In order to assess the performance of the proposed spatial-spectral unmixing model, referred to as SP2U, the unmixing results have been compared to several well-established methods. First, the result of the initialization method has been used as baseline. This  method is conventional~\cite{Bioucas-Dias2010} and consists in extracting endmembers using VCA method~\cite{Nascimento2005} and then solving a fully constrained least square (FCLS) problem. This first method is referred to as by VCA+FCLS hereafter.

The second compared method uses again a FCLS method to estimate the abundance vectors but uses an alternative endmember extraction algorithm. This method, called SISAL~\cite{Bioucas-Dias2009}, tries to estimate the minimum volume simplex containing the observed hyperspectral data by solving a non-convex problem using a splitting augmented Lagrangian technique.

The third compared method relies on a similar linear mixing model assumed by VCA+FCLS and SISAL+FCLS. However, instead of estimating the endmember signatures and abundances sequentially, it performs a joint estimation, yielding a non-negative matrix factorization (NMF) task with an additional sum-to-one constraint. This method referred to as NMF in the sequel, is a depreciated version of the SP2U problem \eqref{eq:cofact-pb} where $\lambda_1=\lambda_2=\lambda_z=0$ and has been solved and initialized similarly.

The fourth method SUnSAL-TV was introduced in~\cite{Iordache2012} and proposes to solve a conventional linear unmixing problem with an additional spatial regularization term to incorporate spatial information. The regularization term is chosen as a total variation applied to the abundance maps $\mathbf{A}$. It promotes in particular similarity of abundance vectors of neighboring pixels. In this case, the local information is used whereas SP2U method relates pixels sharing the same spatial context, akin to a non-local framework. It is important to note that this method does not estimate the endmember matrix which is estimated beforehand using VCA or SISAL. 

The fifth method, denoted n-SP2U, is a naive counterpart of the proposed SP2U method. Instead of using the coupling term introduced in Section \ref{sec:coupling}, the abundance matrix $\mathbf{A}$ and the coding coefficients $\mathbf{U}$ are directly considered equal yielding the following problem
\begin{align}
\label{eq:naive-pb}
  \min_{\mathbf{M},\mathbf{A},\mathbf{D}} &\frac{\lambda_0}{2} \norm{\mathbf{Y} - \mathbf{M} \mathbf{A}}_{\mathrm{F}}^2 + \imath_{\mathbb{R}^{d_1\times R_1}_+}{(\mathbf{M})} \nonumber\\
  &+ \frac{\lambda_1}{2} \norm{\mathbf{S} - \mathbf{D} \mathbf{A}}_{\mathrm{F}}^2 + \imath_{\mathbb{R}^{d_2 \times R_2}_+}{(\mathbf{D})} + \imath_{\mathbb{S}_{R_1}^P}(\mathbf{A}).
\end{align}
This method is considered for comparison since it may seem natural when willing to couple factorizations associated with spatial and spectral unmixing. However, it actually appears very unlikely to perform well in real scenarios. Indeed, the naive model n-SP2U actually enforces a same size of dictionary for the spatial mixing and the spectral mixing. In realistic cases, we expect to have less spectral signatures than spatial signatures. Indeed, a given elementary material can be present with several spatial patterns whereas a given spatial pattern would be unlikely associated to distinct spectral signatures. To account for the case of a given spectral signature associated with several spatial signatures, the n-SP2U model would need a larger endmember matrix $\mathbf{M}$ with several equal columns, each one associated with a specific spatial signature in $\mathbf{D}$. The resulting estimated abundance matrix $\mathbf{A}$ would need to follow a fusion process to sum abundance maps corresponding to the same spectral signatures.

The last compared method, denoted as c-SPU, is another simplified version of SP2U where the spatial data fitting term has been removed. It corresponds to a NMF-based spectral unmixing method combined with a clustering of the spectral abundance vectors. The method c-SPU solves the problem
\begin{align}
\label{eq:cSPU-pb}
  \min_{\mathbf{M},\mathbf{A},\mathbf{B},\mathbf{Z}} &\frac{\lambda_0}{2} \norm{\mathbf{Y} - \mathbf{M} \mathbf{A}}_{\mathrm{F}}^2 + \imath_{\mathbb{R}^{d_1 \times R_1}_+}{(\mathbf{M})} + \imath_{\mathbb{S}_{R_1}^P}(\mathbf{A}) \nonumber\\
  &+ \frac{\lambda_2}{2} \norm{ \mathbf{A} - \mathbf{B} \mathbf{Z}}_{\mathrm{F}}^2 + \frac{\lambda_z}{2} \mathrm{Tr}(\mathbf{Z}^T\mathbf{V}\mathbf{Z}) \nonumber\\
  &+ \imath_{\mathbb{R}^{(R_1+R_2)\times K}_+}{(\mathbf{B})} + \imath_{\mathbb{S}_{K}^P}(\mathbf{Z}).
\end{align}
Comparing the results provided by c-SPU with those obtained by the proposed method SP2U aims at demonstrating the interest of considering the spatial data fitting term. At a lesser extent, it will also measure the benefit of introducing the clustering term, which is expected to act as a regularizer for the spectral abundance matrix $\mathbf{A}$.

\subsection{Performance criteria}
Performance of all methods has been assessed in term of endmember estimation using the average spectral angle mapper (aSAM)
\begin{equation}
\label{eq:aSAM}
  \mathrm{aSAM}(\mathbf{M}) = \frac{1}{R_1} \sum_{r=1}^{R_1} \arccos \left( \frac{ \langle{\mathbf{m}^{(\mathrm{ref})}_r} | \mathbf{m}_r\rangle}{\|{\mathbf{m}^{(\mathrm{ref})}_r}\|_2 
  \|{\mathbf{m}_r}\|_2} \right),
\end{equation}
and also in term of abundance estimation using the root mean square error (RMSE)
\begin{equation}
\label{eq:rmse}
  \mathrm{RMSE}(\mathbf{A}) = \sqrt{\frac{1}{PR_1} \norm{\mathbf{A}^{(\mathrm{ref})}-\mathbf{A}}_{\mathrm{F}}^2},
\end{equation}
where $\mathbf{m}^{(\mathrm{ref})}_r$ and $\mathbf{A}$ are the $r$th actual endmember signature and the actual abundance matrix, respectively.

\begin{table*}
  \centering
  \caption{\textsf{\small Image 1}: quantitative results (averaged over $10$ trials).\label{tab:res-synth-1}}
  \begin{tabular}[ht!]{lcccc}\toprule
    Model & aSAM$({\mathbf{M}})$ & RE & RMSE$({\mathbf{A}})$ & Time (s) \\
    \midrule
    VCA+FCLS      & $0.180$ ($\pm 1.1\times 10^{-2}$) & $6.86\times 10^{-3}$ ($\pm 6.3\times 10^{-3}$) & $0.150$ ($\pm 1.9\times 10^{-2}$) & $19$ ($\pm 11$) \\
    SISAL+FCLS    & $0.151$ ($\pm 3.4\times 10^{-3}$) & $2.81\times 10^{-3}$ ($\pm 3.5\times 10^{-6}$) & $0.114$ ($\pm 3.9\times 10^{-3}$) & $23$ ($\pm 0.1$) \\
    NMF           & $0.175$ ($\pm 5.6\times 10^{-3}$) & $3.86\times 10^{-3}$ ($\pm 9.8\times 10^{-4}$) & $0.151$ ($\pm 2.1\times 10^{-2}$) & $27$ ($\pm 29$) \\
    VCA+SUnSAL-TV     & $0.180$ ($\pm 1.1\times 10^{-2}$) & $7.61\times 10^{-3}$ ($\pm 4.5\times 10^{-3}$) & $0.132$ ($\pm 3.2\times 10^{-2}$) & $27$ ($\pm 0.1$) \\
    SISAL+SUnSAL-TV& $0.151$ ($\pm 2.9\times 10^{-3}$) & $4.6\times 10^{-3}$ ($\pm 1.1\times 10^{-4}$)  & $\mathbf{0.0989}$ ($\pm 4.1\times 10^{-3}$) & $28$ ($\pm 0.3$) \\
    n-SP2U   & $0.188$ ($\pm 1.5\times 10^{-2}$) & $28.1\times 10^{-3}$ ($\pm 1.2\times 10^{-3}$) & $0.192$ ($\pm 9.6\times 10^{-3}$) & $93$ ($\pm 14$) \\
    c-SPU    & $0.178$ ($\pm 3.8\times 10^{-3}$) & $10.9\times 10^{-3}$ ($\pm 8.0\times 10^{-4}$) & $0.139$ ($\pm 9.1\times 10^{-3}$) & $\mathbf{8}$ ($\pm 0.5$) \\
    SP2U         & $\mathbf{0.108}$ ($\pm 2.2\times 10^{-2}$) & $6.88\times 10^{-3}$ ($\pm 3.5\times 10^{-4}$) & $0.166$ ($\pm 7.2\times 10^{-2}$) & $409$ ($\pm 38$) \\
    \bottomrule
  \end{tabular}
\end{table*}

\begin{table*}
  \centering
  \caption{\textsf{\small Image 2}: quantitative results (averaged over $10$ trials).\label{tab:res-synth-2}}
  \begin{tabular}[ht!]{lcccc}\toprule
    Model & aSAM$({\mathbf{M}})$ & RE & RMSE$({\mathbf{A}})$ & Time (s) \\
    \midrule
    VCA+FCLS    & $0.176$ ($\pm 5.8\times 10^{-3}$) & $8.80\times 10^{-3}$ ($\pm 2.2\times 10^{-3}$)   & $0.246$ ($\pm 4.2\times 10^{-3}$) & $100$  ($\pm 27$) \\
    SISAL+FCLS  & $0.187$ ($\pm 1.7\times 10^{-2}$) & $4.61\times 10^{-3}$ ($\pm 5.0\times 10^{-6}$) & $0.145$ ($\pm 2.3\times 10^{-2}$) & $57$   ($\pm 0.5$) \\
    NMF         & $0.178$ ($\pm 5.9\times 10^{-3}$) & $4.87\times 10^{-3}$ ($\pm 6.3\times 10^{-3}$)  & $0.246$ ($\pm 4.2\times 10^{-3}$) & $109$  ($\pm 26$) \\
    VCA+SUnSAL-TV   & $0.176$ ($\pm 5.8\times 10^{-3}$) & $9.48\times 10^{-3}$ ($\pm 6.4\times 10^{-4}$)  & $0.229$ ($\pm 3.6\times 10^{-3}$) & $81$ ($\pm 0.7$) \\
    SISAL+SUnSAL-TV & $0.189$ ($\pm 9.6\times 10^{-3}$) & $4.74\times 10^{-3}$ ($\pm 5.4\times 10^{-5}$)  & $0.131$ ($\pm 1.2\times 10^{-2}$) & $81$ ($\pm 2$) \\
    n-SP2U & $0.190$ ($\pm 1.8\times 10^{-2}$) & $35.3\times 10^{-3}$ ($\pm 4.1\times 10^{-3}$)   & $0.212$ ($\pm 3.0\times 10^{-2}$) & $518$  ($\pm 77$) \\
    c-SPU  & $0.177$ ($\pm 6.1\times 10^{-3}$) & $18.0\times 10^{-3}$ ($\pm 7.6\times 10^{-4}$)   & $0.232$ ($\pm 1.1\times 10^{-2}$) & $\mathbf{44}$  ($\pm 13$) \\
    SP2U       & $\mathbf{0.155}$ ($\pm 1.4\times 10^{-2}$) & $9.74\times 10^{-3}$ ($\pm 4.3\times 10^{-4}$)  & $\mathbf{0.125}$ ($\pm 3.9\times 10^{-2}$) & $1174$ ($\pm 62$) \\
    \bottomrule
  \end{tabular}
\end{table*}

Two additional information have also been included in the results. The processing time, which includes the initialization, the endmember extraction and the abundances estimation; and the reconstruction error which measure how the model fits to the observed data
\begin{equation}
\label{eq:metric}
  \mathrm{RE} = \sqrt{\frac{1}{Pd_1} \norm{\mathbf{Y}-\mathbf{M}\mathbf{A}}_{\mathrm{F}}^2}.
\end{equation}
It is worth noting that RE should not be understood as a criterion of unmixing performance. It rather measures the ability of a given model to fit the observations. Thus a very low RE may not be systematically suitable since it could be explained by overfitting. Conversely, a high RE may help to diagnose a model unreliability or issues in algorithmic convergences. However, when comparing a given set of methods, REs of the same order of magnitude ensure that all methods are able to describe the data with similar accuracy and the criteria for unmixing performance (RMSE and SAM) can be compared fairly.

\subsection{Results}
\label{sec:synth-res}

As stated in Section~\ref{sec:spa-mod}, the spatial feature matrix $\mathbf{S}$ has been extracted from the panchromatic image. For each pixel, the spatial feature vector $\mathbf{s}_p$ ($p\in \left\{1,\ldots,P\right\}$) is obtained by concatenating the values of the pixels in a $11 \times 11$-pixel patch centered on the considered pixel. This choice may seem very naive but patch-based image decompositions have proven their interest for many tasks. This choice has also the advantage of offering a direct interpretation of the spatial content and cluster centroids as small $11$-by-$11$ patches. Besides, designing the most appropriate spatial feature is out of the scope of this paper whose main objective is to introduce the concept of spatial-spectral unmixing. Moreover, for these experiments, the actual number of endmembers has been assumed known and thus $R_1 = 4$ for \textsf{\small Image 1} and $R_1 = 9$ for \textsf{\small Image 2}. The number of dictionary atoms and clusters have been empirically adjusted and set such that $R_2=20$ and $K = 30$ for \textsf{\small Image 1} and $R_2 = 30$ and $K=40$ for \textsf{\small Image 2}. It is worth noting that increasing these two parameters tends to improve the performance up to a certain point where a slow decreasing can be observed. Hence, the choice of these values is not critical as long as they are high enough. It can be explained by the fact that a sufficient number of atoms and centroids is needed to explain the data. However, beyond a certain value, increasing these parameters reduces the regularization induced by the clustering. In a more general case, using features more robust to rotation and translation deformation would likely allow to reduce the number of needed clusters and dictionary atoms. Moreover, the weighting terms of the various methods have been adjusted manually using a gridsearch algorithm in order to obtain consistent results. In particular, weighting coefficients of SP2U method have been set to ${\tilde{\lambda}_0} = {\tilde{\lambda}_1} = \lambda_2 = 1.0$ and $\lambda_z = 0.1$.

\pgfplotstableset{
    create on use/wav/.style={
        create col/copy column from table={Fig/ResSynth/endm_wavelength.txt}{Wavelength}
    }
}

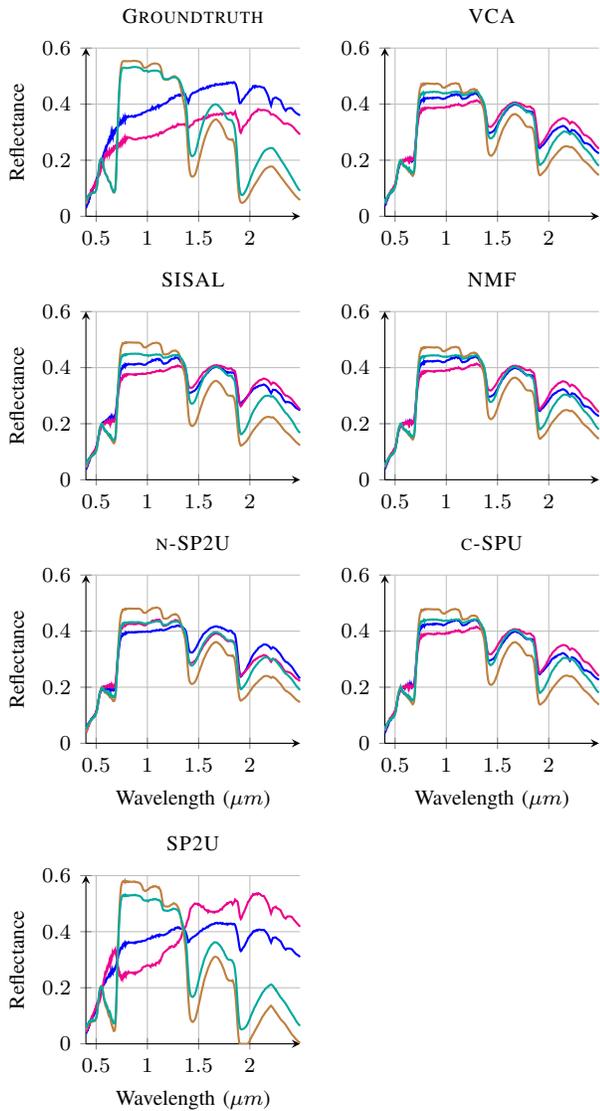
\begin{figure}
  \centering
  \begin{tabular}{@{}cc}
    \begin{tikzpicture}
      \begin{axis}[width=0.5\columnwidth,cycle list name=mycolorlist,ymin=0.,ymax=0.6,grid,axis x line=left,axis y line=left,ylabel={Reflectance},ylabel style={align=center},font=\footnotesize,title=\sc{Groundtruth}]
        \addplot+[thick] table[x=wav,y expr=0.01*\thisrowno{0}] {Fig/ResSynth/endm_true.txt};
        \addplot+[thick] table[x=wav,y expr=0.01*\thisrowno{1}] {Fig/ResSynth/endm_true.txt};
        \addplot+[thick] table[x=wav,y expr=0.01*\thisrowno{2}] {Fig/ResSynth/endm_true.txt};
        \addplot+[thick] table[x=wav,y expr=0.01*\thisrowno{3}] {Fig/ResSynth/endm_true.txt};
      \end{axis};
    \end{tikzpicture}&
    \begin{tikzpicture}
      \begin{axis}[width=0.5\columnwidth,cycle list name=mycolorlist,ymin=0.,ymax=0.6,grid,axis x line=left,axis y line=left,font=\footnotesize,title=\sc{VCA}]
        \addplot+[thick] table[x=wav,y expr=0.01*\thisrowno{0}] {Fig/ResSynth/endm_VCA.txt};
        \addplot+[thick] table[x=wav,y expr=0.01*\thisrowno{1}] {Fig/ResSynth/endm_VCA.txt};
        \addplot+[thick] table[x=wav,y expr=0.01*\thisrowno{2}] {Fig/ResSynth/endm_VCA.txt};
        \addplot+[thick] table[x=wav,y expr=0.01*\thisrowno{3}] {Fig/ResSynth/endm_VCA.txt};
      \end{axis};
    \end{tikzpicture}\\
    \begin{tikzpicture}
      \begin{axis}[width=0.5\columnwidth,cycle list name=mycolorlist,ymin=0.,ymax=0.6,grid,axis x line=left,axis y line=left,ylabel={Reflectance},ylabel style={align=center},font=\footnotesize,title=\sc{SISAL}]
        \addplot+[thick] table[x=wav,y expr=0.01*\thisrowno{0}] {Fig/ResSynth/endm_SISAL.txt};
        \addplot+[thick] table[x=wav,y expr=0.01*\thisrowno{1}] {Fig/ResSynth/endm_SISAL.txt};
        \addplot+[thick] table[x=wav,y expr=0.01*\thisrowno{2}] {Fig/ResSynth/endm_SISAL.txt};
        \addplot+[thick] table[x=wav,y expr=0.01*\thisrowno{3}] {Fig/ResSynth/endm_SISAL.txt};
      \end{axis};
    \end{tikzpicture}&
    \begin{tikzpicture}
      \begin{axis}[width=0.5\columnwidth,cycle list name=mycolorlist,ymin=0.,ymax=0.6,grid,axis x line=left,axis y line=left,font=\footnotesize,title=\sc{NMF}]
        \addplot+[thick] table[x=wav,y expr=0.01*\thisrowno{0}] {Fig/ResSynth/endm_mod0.txt};
        \addplot+[thick] table[x=wav,y expr=0.01*\thisrowno{1}] {Fig/ResSynth/endm_mod0.txt};
        \addplot+[thick] table[x=wav,y expr=0.01*\thisrowno{2}] {Fig/ResSynth/endm_mod0.txt};
        \addplot+[thick] table[x=wav,y expr=0.01*\thisrowno{3}] {Fig/ResSynth/endm_mod0.txt};
      \end{axis};
    \end{tikzpicture}\\
    \begin{tikzpicture}
      \begin{axis}[width=0.5\columnwidth,cycle list name=mycolorlist,ymin=0.,ymax=0.6,grid,axis x line=left,axis y line=left,xlabel={Wavelength ($\mu m$)},ylabel={Reflectance},ylabel style={align=center},font=\footnotesize,title=\sc{n-SP2U}]
        \addplot+[thick] table[x=wav,y expr=0.01*\thisrowno{0}] {Fig/ResSynth/endm_mod1.txt};
        \addplot+[thick] table[x=wav,y expr=0.01*\thisrowno{1}] {Fig/ResSynth/endm_mod1.txt};
        \addplot+[thick] table[x=wav,y expr=0.01*\thisrowno{2}] {Fig/ResSynth/endm_mod1.txt};
        \addplot+[thick] table[x=wav,y expr=0.01*\thisrowno{3}] {Fig/ResSynth/endm_mod1.txt};
      \end{axis};
    \end{tikzpicture}&
    \begin{tikzpicture}
      \begin{axis}[width=0.5\columnwidth,cycle list name=mycolorlist,ymin=0.,ymax=0.6,grid,axis x line=left,axis y line=left,xlabel={Wavelength ($\mu m$)},font=\footnotesize,title=\sc{c-SPU}]
        \addplot+[thick] table[x=wav,y expr=0.01*\thisrowno{0}] {Fig/ResSynth/endm_mod5.txt};
        \addplot+[thick] table[x=wav,y expr=0.01*\thisrowno{1}] {Fig/ResSynth/endm_mod5.txt};
        \addplot+[thick] table[x=wav,y expr=0.01*\thisrowno{2}] {Fig/ResSynth/endm_mod5.txt};
        \addplot+[thick] table[x=wav,y expr=0.01*\thisrowno{3}] {Fig/ResSynth/endm_mod5.txt};
      \end{axis};
    \end{tikzpicture}\\
    \begin{tikzpicture}
      \begin{axis}[width=0.5\columnwidth,cycle list name=mycolorlist,ymin=0.,ymax=0.6,grid,axis x line=left,axis y line=left,xlabel={Wavelength ($\mu m$)},ylabel={Reflectance},ylabel style={align=center},font=\footnotesize,title=\sc{SP2U}]
        \addplot+[thick] table[x=wav,y expr=0.01*\thisrowno{0}] {Fig/ResSynth/endm_mod2.txt};
        \addplot+[thick] table[x=wav,y expr=0.01*\thisrowno{1}] {Fig/ResSynth/endm_mod2.txt};
        \addplot+[thick] table[x=wav,y expr=0.01*\thisrowno{2}] {Fig/ResSynth/endm_mod2.txt};
        \addplot+[thick] table[x=wav,y expr=0.01*\thisrowno{3}] {Fig/ResSynth/endm_mod2.txt};
      \end{axis};
    \end{tikzpicture}&
    \\    
  \end{tabular}
  \caption{\textsf{\small Image 1}: estimated endmembers.\label{fig:res-synth-endm}}
\end{figure}

\begin{figure*}
  \centering
  \footnotesize
  \begin{tabular}{@{}cc@{}c@{}c@{}c@{}c@{}c@{}c@{}c@{}c@{}}
    &\textsc{Groundtruth} & \textsc{SP2U} & \textsc{c-SPU} & \textsc{n-SP2U} & \textsc{SISAL+} & \textsc{VCA+} & \textsc{NMF} & \textsc{SISAL+FCLS} & \textsc{VCA+FCLS} \\
    & & & & & \textsc{SUnSAL-TV} & \textsc{SUnSAL-TV} & & &  \\
    \begin{tikzpicture}
      \node[rectangle,minimum width=0.03\columnwidth,minimum height=0.2\columnwidth] at (0,0 ) {};
      \node[rectangle,fill=blue,minimum width=0.03\columnwidth,minimum height=0.05\columnwidth] at (0,0 ) {};
    \end{tikzpicture}&
    \includegraphics[width=0.2\columnwidth]{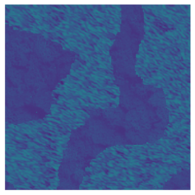}&
    \includegraphics[width=0.2\columnwidth]{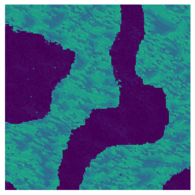}&
    \includegraphics[width=0.2\columnwidth]{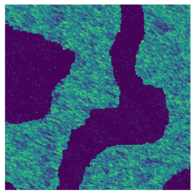}&
    \includegraphics[width=0.2\columnwidth]{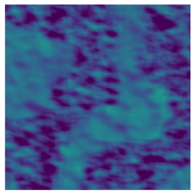}&
    \includegraphics[width=0.2\columnwidth]{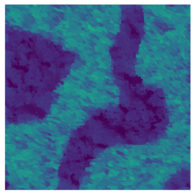}&
    \includegraphics[width=0.2\columnwidth]{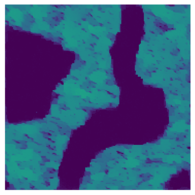}&
    \includegraphics[width=0.2\columnwidth]{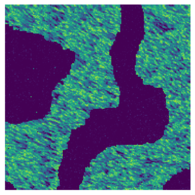}&
    \includegraphics[width=0.2\columnwidth]{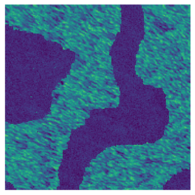}&
    \includegraphics[width=0.2\columnwidth]{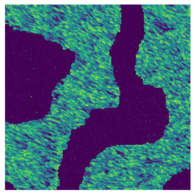}\\
    \begin{tikzpicture}
      \node[rectangle,minimum width=0.03\columnwidth,minimum height=0.2\columnwidth] at (0,0 ) {};
      \node[rectangle,fill=magenta,minimum width=0.03\columnwidth,minimum height=0.05\columnwidth] at (0,0 ) {};
    \end{tikzpicture}&
    \includegraphics[width=0.2\columnwidth]{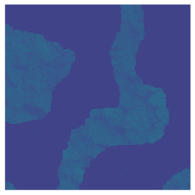}&
    \includegraphics[width=0.2\columnwidth]{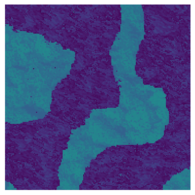}&
    \includegraphics[width=0.2\columnwidth]{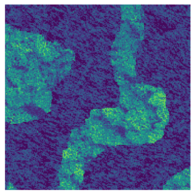}&
    \includegraphics[width=0.2\columnwidth]{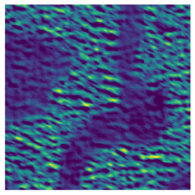}&
    \includegraphics[width=0.2\columnwidth]{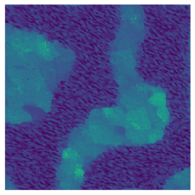}&
    \includegraphics[width=0.2\columnwidth]{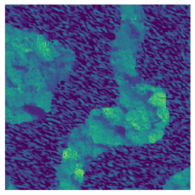}&
    \includegraphics[width=0.2\columnwidth]{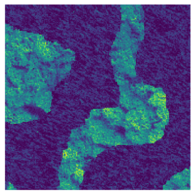}&
    \includegraphics[width=0.2\columnwidth]{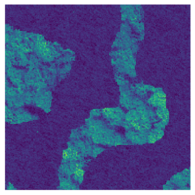}&
    \includegraphics[width=0.2\columnwidth]{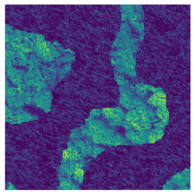}\\
    \begin{tikzpicture}
      \node[rectangle,minimum width=0.03\columnwidth,minimum height=0.2\columnwidth] at (0,0 ) {};
      \node[rectangle,fill=brown,minimum width=0.03\columnwidth,minimum height=0.05\columnwidth] at (0,0 ) {};
    \end{tikzpicture}&
    \includegraphics[width=0.2\columnwidth]{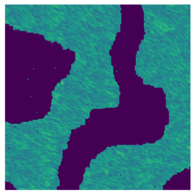}&
    \includegraphics[width=0.2\columnwidth]{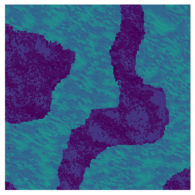}&
    \includegraphics[width=0.2\columnwidth]{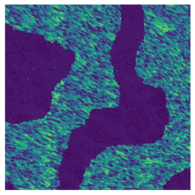}&
    \includegraphics[width=0.2\columnwidth]{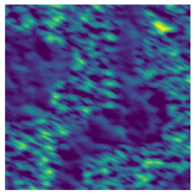}&
    \includegraphics[width=0.2\columnwidth]{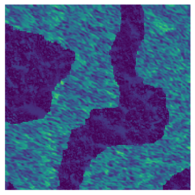}&
    \includegraphics[width=0.2\columnwidth]{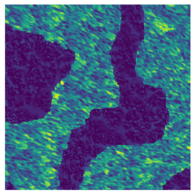}&
    \includegraphics[width=0.2\columnwidth]{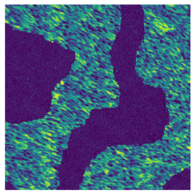}&
    \includegraphics[width=0.2\columnwidth]{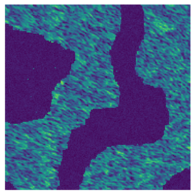}&
    \includegraphics[width=0.2\columnwidth]{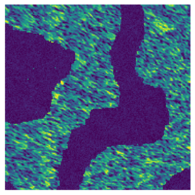}\\
    \begin{tikzpicture}
      \node[rectangle,minimum width=0.03\columnwidth,minimum height=0.2\columnwidth] at (0,0 ) {};
      \node[rectangle,fill=Emerald,minimum width=0.03\columnwidth,minimum height=0.05\columnwidth] at (0,0 ) {};
    \end{tikzpicture}&
    \includegraphics[width=0.2\columnwidth]{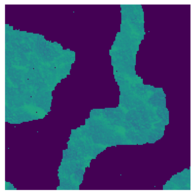}&
    \includegraphics[width=0.2\columnwidth]{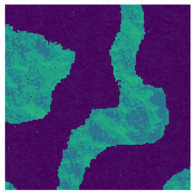}&
    \includegraphics[width=0.2\columnwidth]{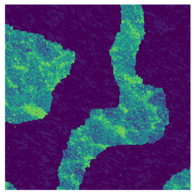}&
    \includegraphics[width=0.2\columnwidth]{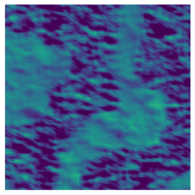}&
    \includegraphics[width=0.2\columnwidth]{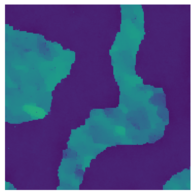}&
    \includegraphics[width=0.2\columnwidth]{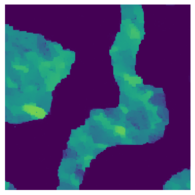}&
    \includegraphics[width=0.2\columnwidth]{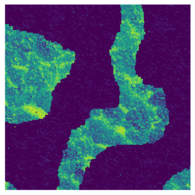}&
    \includegraphics[width=0.2\columnwidth]{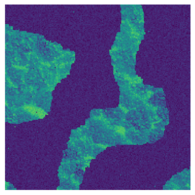}&
    \includegraphics[width=0.2\columnwidth]{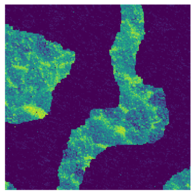}\\
    \multicolumn{10}{c}{
    \begin{tikzpicture}
      \begin{axis}[
        xmin=0,xmax=1,
        ymin=0,ymax=1,
        hide axis,
        scale only axis,
        height=0pt,
        width=0pt,
        colormap/viridis,
        colorbar sampled,
        colorbar horizontal,
        colorbar,
        point meta min=0,
        point meta max=1,
        colorbar style={
          scaled x ticks = false,
          tick label style={/pgf/number format/fixed},
          samples=11,
          height=0.3cm,
          width=8cm,
          xtick={0,0.1,...,1.1},
          /pgf/number format/precision=3
        }]
      \end{axis}
    \end{tikzpicture}}\\
  \end{tabular}
  \caption{\textsf{\small Image 1}: abundance maps (the colored squares refer to the colors used to plot endmembers in Figure~\ref{fig:res-synth-endm}).\label{fig:img1-abund}}
\end{figure*}

As the solution of the considered problem suffers from a permutation ambiguity inherent to factor models, a reordering of the endmembers is thus necessary before any evaluation. In this experiment, this relabeling is performed such that the aSAM is minimum. The quantitative results, averaged over $10$ trials, has then been computed for \textsf{\small Image 1} and \textsf{\small Image 2} and are presented respectively in Tables~\ref{tab:res-synth-1} and~\ref{tab:res-synth-2}. 

The first conclusion of these results is that SP2U method gives the best estimation of the endmember matrix. All other endmember extraction algorithms are clearly behind. In particular, from Figure~\ref{fig:res-synth-endm}, we can see that SP2U is the only method identifying that there are two spectra very different from the others which corresponds to the two soil spectra. Considering the degraded results obtained with c-SPU, it is clear that the spatial model has a real beneficial influence on the results. Another interesting remark is that the NMF model barely improves the initializing point given by VCA+FCLS. It appears to converge in a few iteration to a local minimum close to initialization. Overall, it seems that including the spatial information allows to identify more clearly the endmembers in particular in the considered case where the pure pixel assumption does not hold.

Then, regarding the estimation of abundances, the evaluation is less straightforward since it depends on the estimation of the endmembers. RMSE is computed after the reordering of the endmembers and, for \textsf{\small Image~1}, the best abundance maps are obtained with SISAL+FCLS but they are not associated with the best estimated set of endmembers. The case of \textsf{\small Image~2} is easier to discuss since the best abundance maps, obtained by SP2U, are associated with the best set of endmembers. It is also interesting to consider a qualitative evaluation of the obtained abundance maps depicted in Figure~\ref{fig:img1-abund}. Even if the quantitative results seem to support the quality of the abundance maps retrieved by SUnSAL-TV, the results visually appear overly smooth. On the other hand, abundance maps estimated by SP2U seem visually relevant but the corresponding RMSE suffers from an overestimation of abundances corresponding to soil spectra.

Besides, even if RE is not a relevant criterion to assess the quality of unmixing results, the values reported in Tables~\ref{tab:res-synth-1} and~\ref{tab:res-synth-2} show that most of the models are equally good at recovering mixtures explaining the observed data. The naive counterpart n-SP2U of SP2U exhibits significantly higher REs, which was expected as explained in Section~\ref{sec:comp-meth}. Some methods such as SISAL+FCLS get slightly lower REs. However this can be easily explained by the fact that such a method aims at merely minimizing the RE, which does not necessarily lead to better RMSEs.

Finally, it is interesting to have a look at the computational times. SP2U appears as the slowest method since it inherits from a much richer model. However, the reported computational times should be taken cautiously. Indeed, SUnSAL-TV and SISAL+FCLS were implemented with a fixed number of iterations and are based on Lagrangian augmented splitting methods. Conversely, other methods use a PALM algorithm with a different stopping criterion (see Section~\ref{sec:impl}).

\section{Experiments using real data}
\label{sec:real-exp}

\subsection{Real dataset}
\label{sec:real-data}

\begin{figure}[ht]
  \centering
  \begin{tabular}{@{}cc}
    \includegraphics[width=0.45\columnwidth]{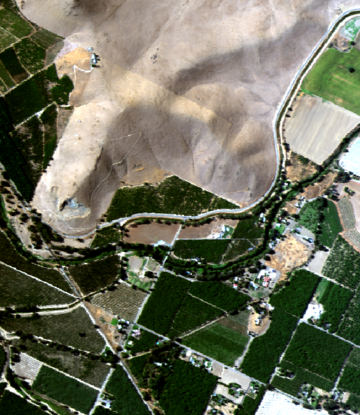}&
    \includegraphics[width=0.45\columnwidth]{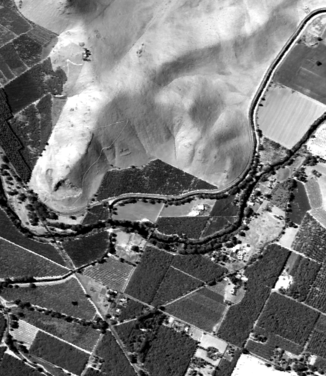}\\
  \end{tabular}
  \caption{AVIRIS image: color composition of hyperspectral image (left) and corresponding panchromatic image (right).\label{fig:data-aviris}}
\end{figure}

The real aerial hyperspectral image used to conduct the following experiment was acquired by AVIRIS in 2013 on a site called Citrus Belt 3, California. The image is composed of $224$ spectral bands from $400$ to $2500$ nanometers with a spatial resolution of $3$m per pixel. After removing bands corresponding to water absorption, a $751 \times 651$-pixel image with $d_1=175$ spectral bands has been finally obtained. A panchromatic image of the scene is computed by normalizing then summing all spectral bands. The resulting image and a color composition of the scene are presented in Figure~\ref{fig:data-aviris}. It is possible to state that the scene includes a desert area and several vegetation areas. Thus several soil and vegetation spectra are expected to be identified.

\subsection{Compared methods}

As explained in Section~\ref{sec:pb-stat}, it is common to consider a sum-to-one constraint for abundance vectors to interpret them as proportion vectors. However, this assumption is not always fulfilled in practical scenarios. In the specific case of the considered AVIRIS image, we decide to drop this constraint due to important illumination variation in the image. For example, the desert area on the upper part of the image is a hill and the spectrum energy is almost doubled on its sunny side. In order to get a well-defined problem after dropping the sum-to-one constraint, it is necessary to introduce a new constraint such that there is no scaling ambiguity between $\mathbf{M}$ and $\mathbf{A}$. The choice has been made to enforce a unit norm of the endmember spectra. Thus, the initial sum-to-one constraint was moved from columns of $\mathbf{A}$ to columns of $\mathbf{M}$. Then, to get abundance maps summing to one, it is possible to normalize the obtained solution a posteriori. Similarly the sum-to-one was removed for SUnSAL-TV, n-SP2U and NMF. Moreover, similarly to the synthetic case, parameters of the problem have been adjusted manually and set to ${\tilde{\lambda}_0} = {\tilde{\lambda}_1} = \lambda_2 = 1$ and $\lambda_z = 0.1$, $R_1 = 6$, $R_2 = 20$ and $K = 30$. 

\subsection{Results}
\label{sec:real-res}

\begin{figure}
  \centering
  \begin{tabular}{@{}cc}
    \begin{tikzpicture}
      \begin{axis}[width=0.6\columnwidth,cycle list name=mycolorlist,grid,axis x line=left,axis y line=left,ymajorticks=false,xlabel={Wavelength ($\mu$m)},font=\footnotesize,title=\sc{VCA},unbounded coords=jump]
        \addplot+[thick] table[x = wav,y expr=0.01*\thisrowno{0}] {Fig/ResAviris/endm_VCA.txt};
        \addplot+[thick] table[x = wav,y expr=0.01*\thisrowno{1}] {Fig/ResAviris/endm_VCA.txt};
        \addplot+[thick] table[x = wav,y expr=0.01*\thisrowno{2}] {Fig/ResAviris/endm_VCA.txt};
        \addplot+[thick] table[x = wav,y expr=0.01*\thisrowno{3}] {Fig/ResAviris/endm_VCA.txt};
        \addplot+[thick] table[x = wav,y expr=0.01*\thisrowno{4}] {Fig/ResAviris/endm_VCA.txt};
        \addplot+[thick] table[x = wav,y expr=0.01*\thisrowno{5}] {Fig/ResAviris/endm_VCA.txt};
      \end{axis};
    \end{tikzpicture}&
    \begin{tikzpicture}
      \begin{axis}[width=0.6\columnwidth,cycle list name=mycolorlist,grid,axis x line=left,axis y line=left,ymajorticks=false,xlabel={Wavelength ($\mu$m)},font=\footnotesize,title=\sc{SISAL},unbounded coords=jump]
        \addplot+[thick] table[x = wav,y expr=0.01*\thisrowno{0}] {Fig/ResAviris/endm_SISAL.txt};
        \addplot+[thick] table[x = wav,y expr=0.01*\thisrowno{1}] {Fig/ResAviris/endm_SISAL.txt};
        \addplot+[thick] table[x = wav,y expr=0.01*\thisrowno{2}] {Fig/ResAviris/endm_SISAL.txt};
        \addplot+[thick] table[x = wav,y expr=0.01*\thisrowno{3}] {Fig/ResAviris/endm_SISAL.txt};
        \addplot+[thick] table[x = wav,y expr=0.01*\thisrowno{4}] {Fig/ResAviris/endm_SISAL.txt};
        \addplot+[thick] table[x = wav,y expr=0.01*\thisrowno{5}] {Fig/ResAviris/endm_SISAL.txt};
      \end{axis};
    \end{tikzpicture}\\
    \begin{tikzpicture}
      \begin{axis}[width=0.6\columnwidth,cycle list name=mycolorlist,grid,axis x line=left,axis y line=left,ymajorticks=false,xlabel={Wavelength ($\mu$m)},font=\footnotesize,title=\sc{NMF},unbounded coords=jump]
        \addplot+[thick] table[x = wav,y expr=0.01*\thisrowno{0}] {Fig/ResAviris/endm_mod0.txt};
        \addplot+[thick] table[x = wav,y expr=0.01*\thisrowno{1}] {Fig/ResAviris/endm_mod0.txt};
        \addplot+[thick] table[x = wav,y expr=0.01*\thisrowno{2}] {Fig/ResAviris/endm_mod0.txt};
        \addplot+[thick] table[x = wav,y expr=0.01*\thisrowno{3}] {Fig/ResAviris/endm_mod0.txt};
        \addplot+[thick] table[x = wav,y expr=0.01*\thisrowno{4}] {Fig/ResAviris/endm_mod0.txt};
        \addplot+[thick] table[x = wav,y expr=0.01*\thisrowno{5}] {Fig/ResAviris/endm_mod0.txt};
      \end{axis};
    \end{tikzpicture}&
    \begin{tikzpicture}
      \begin{axis}[width=0.6\columnwidth,cycle list name=mycolorlist,grid,axis x line=left,axis y line=left,ymajorticks=false,xlabel={Wavelength ($\mu$m)},font=\footnotesize,title=\sc{n-SP2U},unbounded coords=jump]
        \addplot+[thick] table[x = wav,y expr=0.01*\thisrowno{0}] {Fig/ResAviris/endm_mod1.txt};
        \addplot+[thick] table[x = wav,y expr=0.01*\thisrowno{1}] {Fig/ResAviris/endm_mod1.txt};
        \addplot+[thick] table[x = wav,y expr=0.01*\thisrowno{2}] {Fig/ResAviris/endm_mod1.txt};
        \addplot+[thick] table[x = wav,y expr=0.01*\thisrowno{3}] {Fig/ResAviris/endm_mod1.txt};
        \addplot+[thick] table[x = wav,y expr=0.01*\thisrowno{4}] {Fig/ResAviris/endm_mod1.txt};
        \addplot+[thick] table[x = wav,y expr=0.01*\thisrowno{5}] {Fig/ResAviris/endm_mod1.txt};
      \end{axis};
    \end{tikzpicture}\\
    \begin{tikzpicture}
      \begin{axis}[width=0.6\columnwidth,cycle list name=mycolorlist,grid,axis x line=left,axis y line=left,ymajorticks=false,xlabel={Wavelength ($\mu$m)},font=\footnotesize,title=\sc{c-SPU},unbounded coords=jump]
        \addplot+[thick] table[x = wav,y expr=0.01*\thisrowno{0}] {Fig/ResAviris/endm_mod5.txt};
        \addplot+[thick] table[x = wav,y expr=0.01*\thisrowno{1}] {Fig/ResAviris/endm_mod5.txt};
        \addplot+[thick] table[x = wav,y expr=0.01*\thisrowno{2}] {Fig/ResAviris/endm_mod5.txt};
        \addplot+[thick] table[x = wav,y expr=0.01*\thisrowno{3}] {Fig/ResAviris/endm_mod5.txt};
        \addplot+[thick] table[x = wav,y expr=0.01*\thisrowno{4}] {Fig/ResAviris/endm_mod5.txt};
        \addplot+[thick] table[x = wav,y expr=0.01*\thisrowno{5}] {Fig/ResAviris/endm_mod5.txt};
      \end{axis};
    \end{tikzpicture}&
    \begin{tikzpicture}
      \begin{axis}[width=0.6\columnwidth,cycle list name=mycolorlist,grid,axis x line=left,axis y line=left,ymajorticks=false,xlabel={Wavelength ($\mu$m)},font=\footnotesize,title=\sc{SP2U},unbounded coords=jump]
        \addplot+[thick] table[x = wav,y expr=0.01*\thisrowno{0}] {Fig/ResAviris/endm_mod2.txt};
        \addplot+[thick] table[x = wav,y expr=0.01*\thisrowno{1}] {Fig/ResAviris/endm_mod2.txt};
        \addplot+[thick] table[x = wav,y expr=0.01*\thisrowno{2}] {Fig/ResAviris/endm_mod2.txt};
        \addplot+[thick] table[x = wav,y expr=0.01*\thisrowno{3}] {Fig/ResAviris/endm_mod2.txt};
        \addplot+[thick] table[x = wav,y expr=0.01*\thisrowno{4}] {Fig/ResAviris/endm_mod2.txt};
        \addplot+[thick] table[x = wav,y expr=0.01*\thisrowno{5}] {Fig/ResAviris/endm_mod2.txt};
      \end{axis};
    \end{tikzpicture}\\
  \end{tabular}
  \caption{AVIRIS image: estimated endmembers. Note that endmembers estimated by NMF, n-SP2U and SP2U have been normalized to avoid scaling ambiguity intrinsic of the estimation method. \label{fig:res-aviris-endm}}
\end{figure}
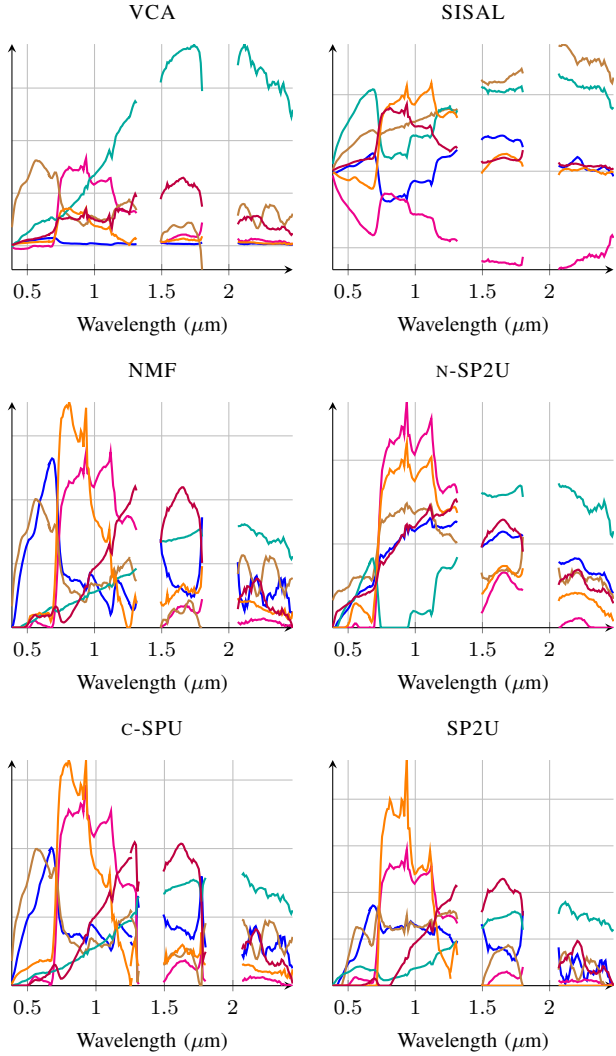

Since no groundtruth is available for this dataset only qualitative evaluations of the various methods are performed. First, Figure~\ref{fig:res-aviris-endm} shows the endmembers estimated by all compared methods. As explained in the previous paragraph, endmembers have been normalized except for SISAL and VCA. Regarding SISAL results, it is possible to note that the method estimates endmember signatures taking negative values. Negative endmembers can not be interpreted as real reflectance spectra and SISAL thus appears the worst compared methods. This method tries to identify a minimum volume simplex containing the observations under the assumption that the observations belong to a $(R_{1}-1)$-dimensional affine set. Thus, these poor results could be explained by a high noise level or non-linear mixtures. It is difficult to objectively compare the results of the other methods. However, the result obtained with SP2U method seems consistent with the visual content of the image since we can clearly identify \emph{i)} two vegetation spectra (plotted in pink and orange) with strong absorbance in the visible domain and strong reflectance in the near-infrared domain~\cite{Myneni1995} \emph{ii)} two soil spectra (plotted in blue and brown) with an increase of the reflectance from $0.4\mu$m to $1\mu$m~\cite{Baumgardner1986}.

\pgfplotstableset{
    create on use/wav/.style={
        create col/copy column from table={Fig/ResAviris/endm_wavelength.txt}{Wavelength}
    }
}

\begin{figure*}
  \centering
  \begin{tabular}{@{}cc@{}c@{}c@{}c@{}c@{}c@{}c@{}c@{}}
    & \textsc{SP2U} & \textsc{c-SPU} & \textsc{n-SP2U} & \textsc{SISAL+} & \textsc{VCA+} & \textsc{NMF} & \textsc{SISAL+FCLS} & \textsc{VCA+FCLS} \\
    & & & & \textsc{SUnSAL-TV} & \textsc{SUnSAL-TV} & & &  \\    
    \begin{tikzpicture}
      \node[rectangle,minimum width=0.05\columnwidth,minimum height=0.25\columnwidth] at (0,0 ) {};
      \node[rectangle,fill=blue,minimum width=0.05\columnwidth,minimum height=0.05\columnwidth] at (0,0 ) {};
    \end{tikzpicture}&
    \includegraphics[width=0.22\columnwidth]{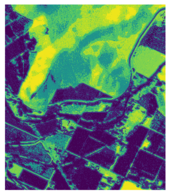}&
    \includegraphics[width=0.22\columnwidth]{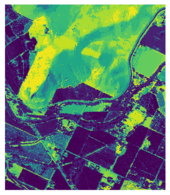}&
    \includegraphics[width=0.22\columnwidth]{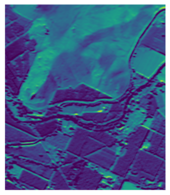}&
    \includegraphics[width=0.22\columnwidth]{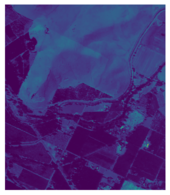}&
    \includegraphics[width=0.22\columnwidth]{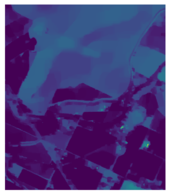}&
    \includegraphics[width=0.22\columnwidth]{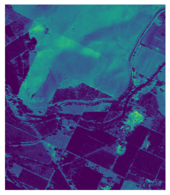}&
    \includegraphics[width=0.22\columnwidth]{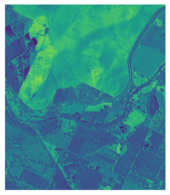}&
    \includegraphics[width=0.22\columnwidth]{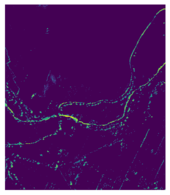}\\
    \begin{tikzpicture}
      \node[rectangle,minimum width=0.05\columnwidth,minimum height=0.25\columnwidth] at (0,0 ) {};
      \node[rectangle,fill=magenta,minimum width=0.05\columnwidth,minimum height=0.05\columnwidth] at (0,0 ) {};
    \end{tikzpicture}&
    \includegraphics[width=0.22\columnwidth]{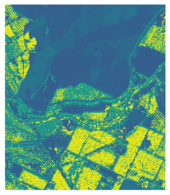}&
    \includegraphics[width=0.22\columnwidth]{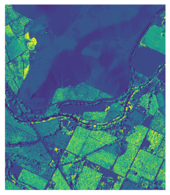}&
    \includegraphics[width=0.22\columnwidth]{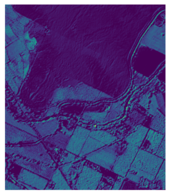}&
    \includegraphics[width=0.22\columnwidth]{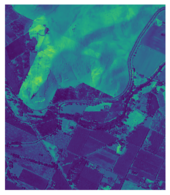}&
    \includegraphics[width=0.22\columnwidth]{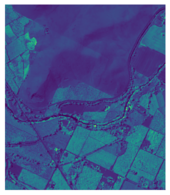}&
    \includegraphics[width=0.22\columnwidth]{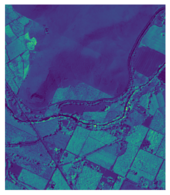}&
    \includegraphics[width=0.22\columnwidth]{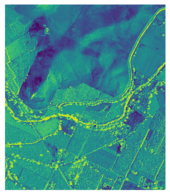}&
    \includegraphics[width=0.22\columnwidth]{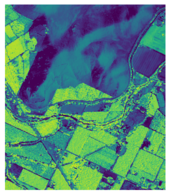}\\
    \begin{tikzpicture}
      \node[rectangle,minimum width=0.05\columnwidth,minimum height=0.25\columnwidth] at (0,0 ) {};
      \node[rectangle,fill=brown,minimum width=0.05\columnwidth,minimum height=0.05\columnwidth] at (0,0 ) {};
    \end{tikzpicture}&
    \includegraphics[width=0.22\columnwidth]{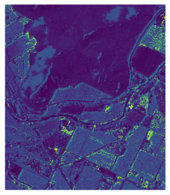}&
    \includegraphics[width=0.22\columnwidth]{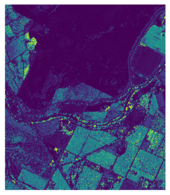}&
    \includegraphics[width=0.22\columnwidth]{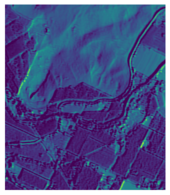}&
    \includegraphics[width=0.22\columnwidth]{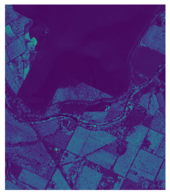}&
    \includegraphics[width=0.22\columnwidth]{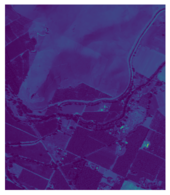}&
    \includegraphics[width=0.22\columnwidth]{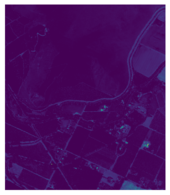}&
    \includegraphics[width=0.22\columnwidth]{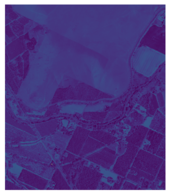}&
    \includegraphics[width=0.22\columnwidth]{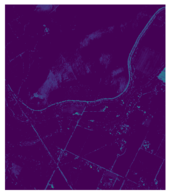}\\
    \begin{tikzpicture}
      \node[rectangle,minimum width=0.05\columnwidth,minimum height=0.25\columnwidth] at (0,0 ) {};
      \node[rectangle,fill=Emerald,minimum width=0.05\columnwidth,minimum height=0.05\columnwidth] at (0,0 ) {};
    \end{tikzpicture}&
    \includegraphics[width=0.22\columnwidth]{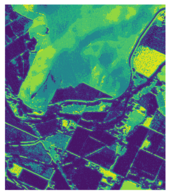}&
    \includegraphics[width=0.22\columnwidth]{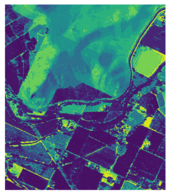}&
    \includegraphics[width=0.22\columnwidth]{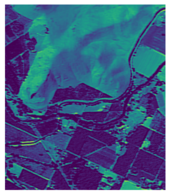}&
    \includegraphics[width=0.22\columnwidth]{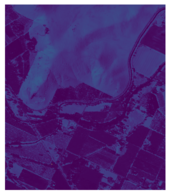}&
    \includegraphics[width=0.22\columnwidth]{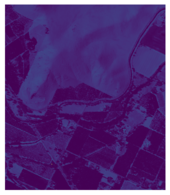}&
    \includegraphics[width=0.22\columnwidth]{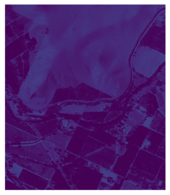}&
    \includegraphics[width=0.22\columnwidth]{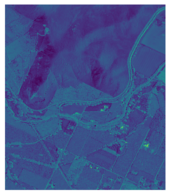}&
    \includegraphics[width=0.22\columnwidth]{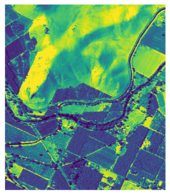}\\
    \begin{tikzpicture}
      \node[rectangle,minimum width=0.05\columnwidth,minimum height=0.25\columnwidth] at (0,0 ) {};
      \node[rectangle,fill=orange,minimum width=0.05\columnwidth,minimum height=0.05\columnwidth] at (0,0 ) {};
    \end{tikzpicture}&
    \includegraphics[width=0.22\columnwidth]{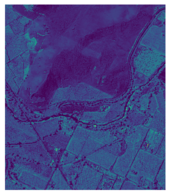}&
    \includegraphics[width=0.22\columnwidth]{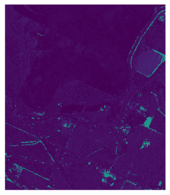}&
    \includegraphics[width=0.22\columnwidth]{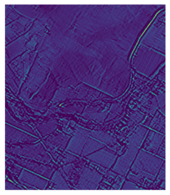}&
    \includegraphics[width=0.22\columnwidth]{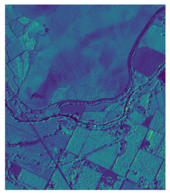}&
    \includegraphics[width=0.22\columnwidth]{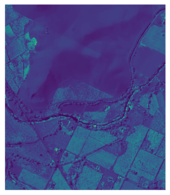}&
    \includegraphics[width=0.22\columnwidth]{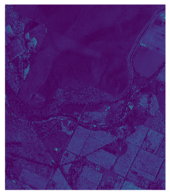}&
    \includegraphics[width=0.22\columnwidth]{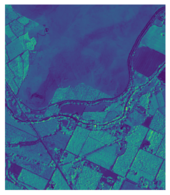}&
    \includegraphics[width=0.22\columnwidth]{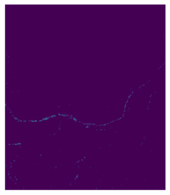}\\
    \begin{tikzpicture}
      \node[rectangle,minimum width=0.05\columnwidth,minimum height=0.25\columnwidth] at (0,0 ) {};
      \node[rectangle,fill=purple,minimum width=0.05\columnwidth,minimum height=0.05\columnwidth] at (0,0 ) {};
    \end{tikzpicture}&
    \includegraphics[width=0.22\columnwidth]{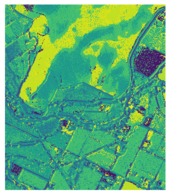}&
    \includegraphics[width=0.22\columnwidth]{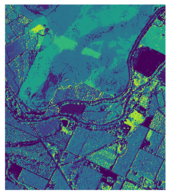}&
    \includegraphics[width=0.22\columnwidth]{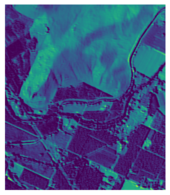}&
    \includegraphics[width=0.22\columnwidth]{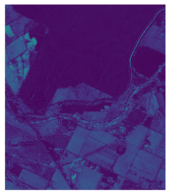}&
    \includegraphics[width=0.22\columnwidth]{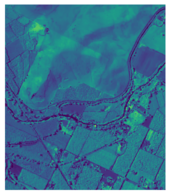}&
    \includegraphics[width=0.22\columnwidth]{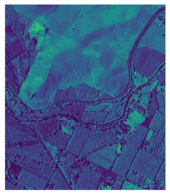}&
    \includegraphics[width=0.22\columnwidth]{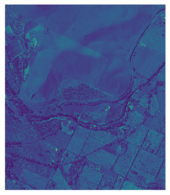}&
    \includegraphics[width=0.22\columnwidth]{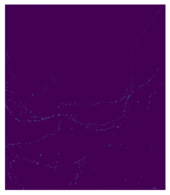}\\
  \end{tabular}
  \caption{AVIRIS image: estimated abundance maps. The colored squares refer to the colors used to plot endmembers in Figure~\ref{fig:res-aviris-endm}. However, no reordering has been performed, i.e., endmembers have no particular relationship between methods. \label{fig:aviris-abund}}
\end{figure*}

Regarding the abundance maps presented in Figure~\ref{fig:aviris-abund}, it seems again that the maps produced by SP2U are consistent with the actual content of the scene. They are in particular spatially consistent with natural edges in the image. Additionally, SP2U results seem to be sparse in the sense that only a few endmembers are used for a given pixel while other methods recover very similar abundance maps with all endmembers, see, e.g., VCA+SUnSAL-TV. From Table~\ref{tab:res-aviris}, it seems that ensuring the sum-to-one constraint makes more difficult to fit to the observations since VCA+FCLS has the highest RE. And, again as expected, SP2U method remains the slowest due to the overload of data to manipulate. 

\begin{table}[ht]
  \centering
  \caption{AVIRIS image: quantitative results.\label{tab:res-aviris}}
  \begin{tabular}[ht!]{lcccc}\toprule
    Model &  RE & Time (s) \\
    \midrule
    VCA+FCLS        & $2.8\times 10^{-3}$  & $12$   \\
    SISAL+FCLS      & $0.14\times 10^{-3}$ & $214$  \\
    NMF             & $0.13\times 10^{-3}$ & $2054$ \\
    VCA+SUnSAL-TV   & $0.88\times 10^{-3}$ & $471$  \\
    SISAL+SUnSAL-TV & $0.15\times 10^{-3}$ & $455$  \\
    n-SP2U          & $1.1\times 10^{-3}$  & $1347$ \\
    c-SPU           & $0.69\times 10^{-3}$ & $578$ \\
    SP2U            & $1.4\times 10^{-3}$  & $7162$ \\
    \bottomrule
  \end{tabular}
\end{table}

\begin{figure}
  \centering
  \begin{tabular}{@{}ccc}
    \textsc{Clusters} &
    \textsc{Spatial} &
    \textsc{Spectral} \\
    \textsc{positioning} &
    \textsc{signatures} &
    \textsc{signatures} \\
    \includegraphics[width=0.3\columnwidth]{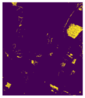}&
    \begin{tikzpicture}
      \node (A) at (0,0 ) {\includegraphics[width=0.15\columnwidth]{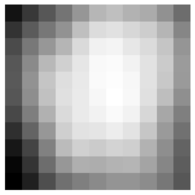}};
      \node[rectangle,minimum width=0.15\columnwidth,minimum height=0.3\columnwidth] at (0,0 ) {};
    \end{tikzpicture}&
    \begin{tikzpicture}
      \begin{axis}[width=0.5\columnwidth,ymin=0.,ymax=4000,axis y line=left,grid,axis x line=left,font=\footnotesize,axis y line=left,ymajorticks=false,xlabel={Wavelength ($\mu$m)},ylabel={Reflectance},ylabel style={align=center},unbounded coords=jump]
        \addplot+[thick] table[x = wav,y expr=\thisrowno{26}] {Fig/ResAviris/spect_clust_mod2_AvirisRoi2.txt};
      \end{axis};
    \end{tikzpicture}\\
    \includegraphics[width=0.3\columnwidth]{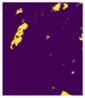}&
    \begin{tikzpicture}
      \node (A) at (0,0 ) {\includegraphics[width=0.15\columnwidth]{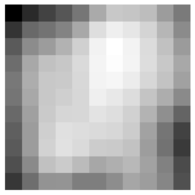}};
      \node[rectangle,minimum width=0.15\columnwidth,minimum height=0.3\columnwidth] at (0,0 ) {};
    \end{tikzpicture}&
    \begin{tikzpicture}
      \begin{axis}[width=0.5\columnwidth,ymin=0.,ymax=4000,axis y line=left,grid,axis x line=left,font=\footnotesize,axis y line=left,ymajorticks=false,xlabel={Wavelength ($\mu$m)},ylabel={Reflectance},ylabel style={align=center},unbounded coords=jump]
        \addplot+[thick] table[x = wav,y expr=\thisrowno{8}] {Fig/ResAviris/spect_clust_mod2_AvirisRoi2.txt};
      \end{axis};
    \end{tikzpicture}\\
    \includegraphics[width=0.3\columnwidth]{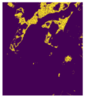}&
    \begin{tikzpicture}
      \node (A) at (0,0 ) {\includegraphics[width=0.15\columnwidth]{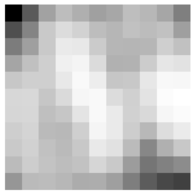}};
      \node[rectangle,minimum width=0.15\columnwidth,minimum height=0.3\columnwidth] at (0,0 ) {};
    \end{tikzpicture}&
    \begin{tikzpicture}
      \begin{axis}[width=0.5\columnwidth,ymin=0.,ymax=4000,axis y line=left,grid,axis x line=left,font=\footnotesize,axis y line=left,ymajorticks=false,xlabel={Wavelength ($\mu$m)},ylabel={Reflectance},ylabel style={align=center},unbounded coords=jump]
        \addplot+[thick] table[x = wav,y expr=\thisrowno{9}] {Fig/ResAviris/spect_clust_mod2_AvirisRoi2.txt};
      \end{axis};
    \end{tikzpicture}\\
    \includegraphics[width=0.3\columnwidth]{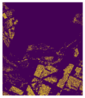}&
    \begin{tikzpicture}
      \node (A) at (0,0 ) {\includegraphics[width=0.15\columnwidth]{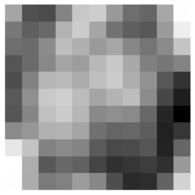}};
      \node[rectangle,minimum width=0.15\columnwidth,minimum height=0.3\columnwidth] at (0,0 ) {};
    \end{tikzpicture}&
    \begin{tikzpicture}
      \begin{axis}[width=0.5\columnwidth,ymin=0.,ymax=4000,axis y line=left,grid,axis x line=left,font=\footnotesize,axis y line=left,ymajorticks=false,xlabel={Wavelength ($\mu$m)},ylabel={Reflectance},ylabel style={align=center},unbounded coords=jump]
        \addplot+[thick] table[x = wav,y expr=\thisrowno{16}] {Fig/ResAviris/spect_clust_mod2_AvirisRoi2.txt};
      \end{axis};
    \end{tikzpicture}\\
    \includegraphics[width=0.3\columnwidth]{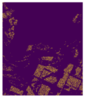}&
    \begin{tikzpicture}
      \node (A) at (0,0 ) {\includegraphics[width=0.15\columnwidth]{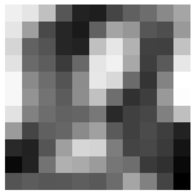}};
      \node[rectangle,minimum width=0.15\columnwidth,minimum height=0.3\columnwidth] at (0,0 ) {};
    \end{tikzpicture}&
    \begin{tikzpicture}
      \begin{axis}[width=0.5\columnwidth,ymin=0.,ymax=4000,axis y line=left,grid,axis x line=left,font=\footnotesize,axis y line=left,ymajorticks=false,xlabel={Wavelength ($\mu$m)},ylabel={Reflectance},ylabel style={align=center},unbounded coords=jump]
        \addplot+[thick] table[x = wav,y expr=\thisrowno{5}] {Fig/ResAviris/spect_clust_mod2_AvirisRoi2.txt};
      \end{axis};
    \end{tikzpicture}\\
  \end{tabular}
  \caption{AVIRIS image: 5 particular clusters described by their spatial positioning (left), mean spatial signature (middle) and mean spectral signature (right).\label{fig:res-aviris-clust}}
\end{figure}

Besides, SP2U is not uniquely a spectral unmixing method and provides much richer interpretation. In Figure~\ref{fig:res-aviris-clust}, the results of the clustering performed by the coupling term are displayed. In particular, this figure shows the spatial position of the clusters, the mean spatial and spectral signatures characterizing the clusters, obtained following \eqref{eq:mean_clusters}. In this example, the first three clusters correspond to soil areas whereas the last two are vegetation, more precisely trees. For instance, the recovered spatial patterns associated with soil are smoother when the wooded areas are characterized by variations of higher frequencies.

\section{Conclusion and perspectives}
\label{sec:ccl}

This paper proposed a new model to interpret hyperspectral images. This method enriched the traditional spectral unmixing modeling by incorporating a spatial analysis of the data. Two data fitting terms, bringing respectively spectral and spatial information, were considered jointly, yielding a spatial-spectral unmixing. This coupled learning process was made possible by the introduction of a clustering-driven coupling term linking the two coding matrices. This clustering process identified groups of pixels with similar spectral and spatial behaviors. 

The experiments conducted on synthetic and real data showed that the proposed method performed very well both at identifying endmembers and estimating abundances. Moreover the relevance of this method was not limited to the unmixing results since the outputs of the clustering task were also of high interest. The identified clusters were characterized by their average spectral signature and spatial context.

It is worth noting that, in this work, the spatial features were merely chosen as elementary patches directly extracted from a virtual panchromatic image generated from the hyperspectral image. This choice allowed a linear approximation of these features to be motivated, exploiting a well-admitted property of image self-similarity. Since the main objective of this work was to introduce the paradigm of
spatial-spectral unmixing, designing the best spatial feature was out of the scope of this paper. However, to further explore the relevance of the proposed model, future works should investigate the benefit of using more complex spatial features. For example, resorting to a convolutional representation of the image may be of high interest to identify shift-invariant textured spatial signatures \cite{Bristow2013,Chabiron2015}.

\appendix
\label{sec:app-optim}
This appendix provides some details regarding the optimization schemes instanced for the proposed cofactorization model. 
Using notations adopted in Section~\ref{sec:optim}, the smooth coupling term can be expressed as
\begin{align*}
g(\mathbf{M},\mathbf{A},\mathbf{D},\mathbf{U},\mathbf{B},\mathbf{Z}) = \frac{\lambda_0}{2} \norm{\mathbf{Y} - \mathbf{M} \mathbf{A}}_{\mathrm{F}}^2 + \frac{\lambda_1}{2} \norm{\mathbf{S} - \mathbf{D} \mathbf{U}}_{\mathrm{F}}^2& \\
+ \frac{\lambda_2}{2} \norm{\left( {\begin{array}{c} \mathbf{A}\\ \mathbf{U}\\\end{array}} \right) - \mathbf{B} \mathbf{Z}}_{\mathrm{F}}^2 + \frac{\lambda_z}{2} \mathrm{Tr}(\mathbf{Z}^T\mathbf{V}\mathbf{Z})&.
\end{align*}

For a practical implementation of PALM, the partial gradients of $g(\cdot)$ and their Lipschitz moduli need to be computed to perform the gradient descent. They are given by
\begin{align*}
\nabla_\mathbf{M} g(\mathbf{M},\mathbf{A},\mathbf{D},\mathbf{U},\mathbf{B},\mathbf{Z}) &= \lambda_0 (\mathbf{M}\mathbf{A}\mathbf{A}^T - \mathbf{Y}\mathbf{A}^T), \\
\nabla_\mathbf{A} g(\mathbf{M},\mathbf{A},\mathbf{D},\mathbf{U},\mathbf{B},\mathbf{Z}) &= \lambda_0 (\mathbf{M}^T\mathbf{M}\mathbf{A} - \mathbf{M}^T\mathbf{Y}) \\ &+ \lambda_2 (\mathbf{A} - \mathbf{B}_1\mathbf{Z}), \\
\nabla_\mathbf{D} g(\mathbf{M},\mathbf{A},\mathbf{D},\mathbf{U},\mathbf{B},\mathbf{Z}) &= \lambda_1 (\mathbf{D}\mathbf{U}\mathbf{U}^T - \mathbf{S}\mathbf{U}^T), \\
\nabla_\mathbf{U} g(\mathbf{M},\mathbf{A},\mathbf{D},\mathbf{U},\mathbf{B},\mathbf{Z}) &= \lambda_1 (\mathbf{D}^T\mathbf{D}\mathbf{U} - \mathbf{D}^T\mathbf{S}) \\ &+ \lambda_2 (\mathbf{U} - \mathbf{B}_2\mathbf{Z}), \\
\nabla_\mathbf{B} g(\mathbf{M},\mathbf{A},\mathbf{D},\mathbf{U},\mathbf{B},\mathbf{Z}) &= \lambda_2 (\mathbf{B}\mathbf{Z}\mathbf{Z}^T - \left( {\begin{array}{c} \mathbf{A}\\ \mathbf{U}\\\end{array}} \right)\mathbf{Z}^T), \\
\nabla_\mathbf{Z} g(\mathbf{M},\mathbf{A},\mathbf{D},\mathbf{U},\mathbf{B},\mathbf{Z}) &= \lambda_2 (\mathbf{B}^T\mathbf{B}\mathbf{Z} - \mathbf{B}^T\left( {\begin{array}{c} \mathbf{A}\\ \mathbf{U}\\\end{array}} \right)) \\ &+ \lambda_z \mathbf{V}\mathbf{Z}
\end{align*}
where $\mathbf{B}_1$ and $\mathbf{B}_2$ correspond to the submatrices of $\mathbf{B}$ defined by the $R_1$ first rows and $R_2$ last rows, respectively, such that ${\mathbf{B} = \left( {\begin{array}{c} \mathbf{B}_1\\ \mathbf{B}_2\\\end{array}} \right)}$.

All partial gradients are globally Lipschitz as functions of the corresponding partial variables. The following Lipschitz moduli can be explicitly derived as
\begin{align}
L_\mathbf{A}(\mathbf{M}) &= \norm{\lambda_0 \mathbf{M}^T\mathbf{M} + \lambda_2 \mathbf{I}_{R_1}}, \nonumber\\
L_\mathbf{M}(\mathbf{A}) &= \norm{\lambda_0 \mathbf{A}\mathbf{A}^T}, \nonumber\\
L_\mathbf{U}(\mathbf{D}) &= \norm{\lambda_1 \mathbf{D}^T\mathbf{D} + \lambda_2 \mathbf{I}_{R_2}}, \nonumber\\
L_\mathbf{D}(\mathbf{U}) &= \norm{\lambda_1 \mathbf{U}\mathbf{U}^T}, \nonumber\\
L_\mathbf{B}(\mathbf{Z}) &= \norm{\lambda_2 \mathbf{Z}\mathbf{Z}^T}, \nonumber\\
L_\mathbf{Z}(\mathbf{B}) &= \norm{\lambda_2 \mathbf{B}^T\mathbf{B} + \lambda_z \mathbf{V}}.
\end{align}

\bibliographystyle{IEEEtran}
\bibliography{strings_all_ref,biblio}

\begin{IEEEbiographynophoto}{Adrien Lagrange}
He received an Engineering degree in Robotics and Embedded Systems from ENSTA ParisTech, France, and the M.Sc. degree in Machine Learning from the Paris Saclay University, both in 2016.
He is currently a Ph.D. student at the National Polytechnic Institute of Toulouse. He is working on the subject of spectral unmixing and classification of hyperspectral images under the supervision of Nicolas Dobigeon and Mathieu Fauvel.
\end{IEEEbiographynophoto}

\begin{IEEEbiographynophoto}{Mathieu Fauvel}
He received the Ph.D. degrees in image and signal processing from the Grenoble Institut of Technology in 2007. From  2008 to 2010, he  was a postdoctoral researcher with the MISTIS Team of the National Institute for Research in Computer Science and Control (INRIA). Since 2011, Dr. Fauvel has been an Associate Professor with the National Polytechnic Institute of Toulouse within the DYNAFOR lab (INRA). His research interests are remote sensing, pattern recognition, and image processing.
\end{IEEEbiographynophoto}

\begin{IEEEbiographynophoto}{Stéphane May}
He received an Engineering degree France in Telecommunications from National Institut of Telecommunications (Evry, France), in 1997. He is currently with the Centre National d'Études Spatiales (French Space Agency), Toulouse, France, where he is developing image processing algorithms and softwares for the exploitation of Earth observation images.
\end{IEEEbiographynophoto}

\begin{IEEEbiographynophoto}{Nicolas Dobigeon}
He received the  Ph.D. degree in Signal Processing from the National Polytechnic Institute of Toulouse in 2012. He was a postdoctoral researcher with the Department of Electrical Engineering and Computer Science, University of Michigan (USA), from 2007 to 2008. Since 2008, he has been with the National Polytechnic Institute of Toulouse, currently with a Professor position. His research interests include statistical signal and image processing.
\end{IEEEbiographynophoto}

\end{document}